\documentclass[lettersize,journal]{IEEEtran}
\IEEEoverridecommandlockouts

\usepackage{cite}
\usepackage{amsmath,amssymb,amsfonts}
\usepackage{graphicx}

\usepackage{dirtytalk}
\usepackage{adjustbox}
\usepackage{multirow}
\usepackage{array}
\usepackage{caption}
\usepackage{booktabs}
\usepackage{lipsum}
\usepackage{amssymb}
\usepackage{amsmath}
\usepackage{bm}
\usepackage{subfig}  
\usepackage{enumitem}
\usepackage{algorithm}
\usepackage{algpseudocode}
\usepackage{subfig} 
\def\BibTeX{{\rm B\kern-.05em{\sc i\kern-.025em b}\kern-.08em
    T\kern-.1667em\lower.7ex\hbox{E}\kern-.125emX}}
\usepackage{balance}
\usepackage{authblk}

\begin{document}
\title{Optimizing Multitask Industrial Processes with Predictive Action Guidance}

\author[1,2]{Naval Kishore Mehta}
\author[1,2]{Arvind}
\author[1,2]{Shyam Sunder Prasad}
\author[1,2]{Sumeet Saurav}
\author[1,2]{Sanjay Singh}
\affil[1]{CSIR-Central Electronics Engineering Research Institute (CSIR-CEERI), India}
\affil[2]{Academy of Scientific and Innovative Research (AcSIR), India}


\maketitle

\begin{abstract}
Monitoring complex assembly processes is critical for maintaining productivity and ensuring compliance with assembly standards. However, variability in human actions and subjective task preferences complicate accurate task anticipation and guidance. To address these challenges, we introduce the Multi-Modal Transformer Fusion and Recurrent Units (MMTF-RU) Network for egocentric activity anticipation, utilizing multi-modal fusion to improve prediction accuracy. Integrated with the Operator Action Monitoring Unit (OAMU), the system provides proactive operator guidance, preventing deviations in the assembly process. OAMU employs two strategies: (1) Top-5 MMTF-RU predictions, combined with a reference graph and an action dictionary, for next-step recommendations; and (2) Top-1 MMTF-RU predictions, integrated with a reference graph, for detecting sequence deviations and predicting anomaly scores via an entropy-informed confidence mechanism. We also introduce Time-Weighted Sequence Accuracy (TWSA) to evaluate operator efficiency and ensure timely task completion. Our approach is validated on the industrial Meccano dataset and the large-scale EPIC-Kitchens-55 dataset, demonstrating its effectiveness in dynamic environments.

\end{abstract}


\begin{figure}
\centering
\includegraphics[width=0.46\textwidth]{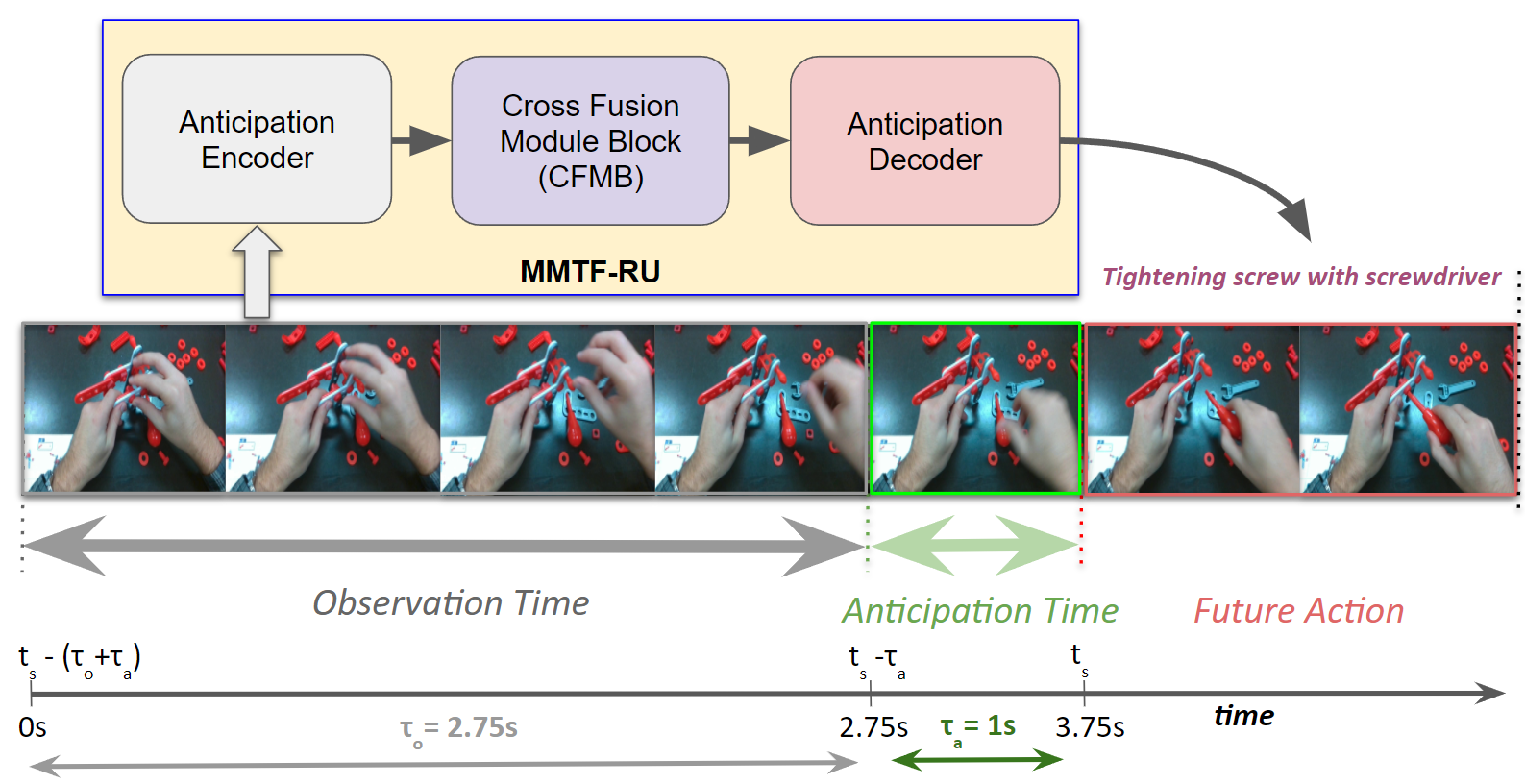}
\caption{\textbf{Egocentric activity anticipation:} Predicting future actions using the MMTF-RU framework, which determines the next action start time $t_s$ after an anticipation interval $\tau_a$, based on the observation time $\tau_o$.}
 \label{overview}
\end{figure}

\section{Introduction}
\label{sec:introduction}
The advent of Industry 5.0 marks a transformative shift toward integrating advanced technologies with human-centric solutions, creating smarter and more responsive industrial environments. The focus is no longer just on humans and robots coexisting but on active collaboration between them to complete tasks more efficiently and with improved outcomes~\cite{raessa2020human,zhang2022human,liu2021unified}. As smart manufacturing advances, dashboard systems, encompassing decision support, data analytics, and human–machine interfaces (HMI), have become essential for monitoring production, analyzing trends, and optimizing key performance indicators. However, as assembly tasks grow more complex and diverse, full automation remains challenging due to the limitations of machines in dexterity and decision-making. Adding to this complexity is the unpredictability of human behavior on the assembly line, which further complicates workflow management~\cite{schirmer2023anomaly}. Furthermore, manual assembly not only introduces risks such as incorrect assembly, missing components, or assembling parts in the wrong sequence, but is also highly influenced by performance-shaping factors like task repetitiveness, increasing product variety, and operator skills or experience. These issues are typically detected post-process during inspections, which limits the ability to prevent errors in real time~\cite{dalle2016integrated}. Monitoring the entire assembly process and intervening promptly to stop mis-assemblies is therefore critical to ensuring product quality.

Anticipating egocentric human intentions is crucial for maintaining smooth operations and ensuring safety in dynamic environments. Robust action anticipation algorithms are key for intelligent systems to plan effectively, enhancing workflow efficiency and safety~\cite{ref18}. By predicting the next steps, systems can provide proactive guidance, prevent anomalies, and alert operators to missing actions or potential dangers, aligning with Industry 5.0’s goal of creating more intelligent, responsive industrial processes~\cite{ref22,ref23,mehta2024iar}.

However, predicting egocentric activities presents challenges like semantic gaps, incomplete observations, ego-motion, and cluttered backgrounds. These issues are further complicated by visual disparities, logical disconnects, and behavioral variability~\cite{ref21}. To address this, various models focus on capturing temporal patterns from past data~\cite{ref11,ref23}, treating action anticipation as a sequence-to-sequence problem using convolutional and recurrent networks~\cite{ref7,ref10,ref18}. Despite progress, challenges remain under strict benchmarks~\cite{ref2,ref5,ref7,ref9,ref11}, compounded by the complexity of human actions, environmental variability, and real-time processing demands~\cite{mehta2024df,ref6}.

\begin{figure}[ht]
\centerline{\includegraphics[scale=.55]{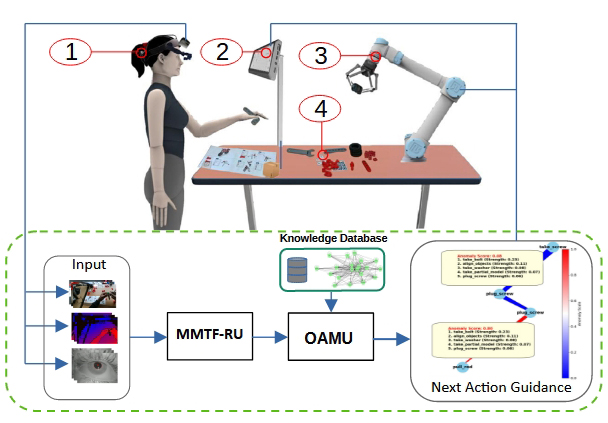}}
\caption{Overview of the collaborative assembly workspace. The setup includes (1) an operator's egocentric view and gaze input to the MMTF-RU model, (2) real-time visual feedback for guidance and anomaly alerts, (3) a robotic arm assisting with tasks, and (4) tools and components on the workbench. The MMTF-RU, integrated with OAMU and a knowledge base, provides next-action guidance for efficient assembly operations.}
\label{fig:collab_workspace}
\end{figure}

In this work, we address the challenges of predicting egocentric activities and modeling complex assembly tasks in dynamic, non-linear industrial environments. To improve action anticipation, we introduce the MMTF-RU model, which leverages multi-modal fusion through a transformer-based encoder and a Cross-Modality Fusion Block (CMFB) to process diverse data streams , as illustrated in Fig.~\ref{overview}. The GRU-based decoder then predicts future actions with anticipation time ($\tau_a$). For handling complex assembly tasks, we introduce the OAMU, which applies a Markov chain-based model to capture and predict workflow transitions using training knowledge database insights. By integrating MMTF-RU’s predictions with OAMU's sequence modeling, the system provides recommendations (tools, actions, or verb-noun combinations) and proactively detects anomalies, ensuring adaptability and precision in dynamic environments, as depicted in Fig.~\ref{fig:collab_workspace}. Time-Weighted Sequence Accuracy (TWSA) evaluates operator efficiency, identifies bottlenecks, and optimizes task execution within ideal time frames. This approach validated on two prominent benchmarks: the Meccano~\cite{ref5} dataset for industrial assembly tasks, and the EPIC-Kitchens-55~\cite{ref6} dataset for egocentric vision and daily activity anticipation. To our knowledge, this is the first work to combine the problems of action anticipation with human activity evaluation in a multi-task industrial setting. The key contributions of this work are as follows:
\begin{itemize}
[topsep=2pt, itemsep=3pt, parsep=3pt]
\item We propose the MMTF-RU model for egocentric activity anticipation, featuring a transformer-based encoder, CMFB for pairwise modality fusion, and a GRU-based decoder.

\item Our approach delivers  state-of-the-art performance in action, verb, and noun anticipation on the industrial Meccano dataset~\cite{ref5}, while demonstrating competitive results across the same tasks on the EPIC-Kitchens-55~\cite{ref6} dataset.

\item  We introduce the OAMU, integrated with MMTF-RU, to recommend next actions and prevent anomalies in dynamic industrial environments using sequence transitions and an entropy-informed confidence mechanism.

\item  We propose the TWSA metric for evaluating operator efficiency in complex assembly tasks, offering targeted insights for optimization and pinpointing critical improvement areas.

\end{itemize} 
The remaining structure of this paper is as follows: In Section \ref{RW}, we provide a brief overview of recent methods in egocentric activity anticipation and explore worker assistance systems. Section \ref{PA} presents an overview of our proposed model and the framework for real-time operator guidance. In Section \ref{CD}, we describe the Meccano~\cite{ref5} and EPIC-Kitchens-55~\cite{ref6} datasets, along with the implementation details used in our experiments. We also present the evaluation results, including a comparison of model performance across both datasets with existing methods, as well as an evaluation of the framework's effectiveness in operator guidance, anomaly prevention, and task efficiency. Finally, Section \ref{CN} summarizes the key findings and outlines directions for future research.

\section{Related Work}
\label{RW}

This section reviews relevant research in two areas. Section \ref{Egocentric} covers egocentric activity anticipation methods, while Section  \ref{WASystem} examines worker assistance systems in manufacturing, with a focus on their role in enhancing productivity and minimizing errors.

\subsection{Human Activity Anticipation} 
\label{Egocentric}

Anticipating human activities entails recognizing both interacted objects and patterns of target actions in the immediate future. Anticipating human actions has gained interest in the research community HRC, with applications in manufacturing, health care, smart home, etc. Wang et al.~\cite{wang2021predicting} developed a multimodal learning method for robots to anticipate human handover intentions using natural language, EMG, and IMU sensors with extreme learning machines, predicting commands like stop, continue, or slow down. While this framework enables action anticipation, it is limited by sensor data quality and adaptability to unstructured environments. Wong et al.~\cite{wong2023vision} developed a multimodal method to distinguish between intentional and unintentional interactions with collaborative robots using touch, body pose, and head gaze, enabling real-time anticipation of user actions. However, reliance on touch data may limit its effectiveness in scenarios with minimal or indirect physical contact. The egocentric perspective serves as a rich source of information regarding the user's intentions. This predictive capability proves particularly valuable in complex scenarios involving human interactions, either with other individuals or machines. In these scenarios, the intelligent system or agent operates as an assistant, providing advance guidance that is both timely and well-grounded, utilizing insights derived from the user's egocentric perspective~\cite{ref22,ref23}. Approaches to action anticipation can be broadly classified into LSTM/RNN-based methods~\cite{ref7,ref9,ref44}, transformer-based methods focusing on feature learning ~\cite{ref45,ref47} and temporal modeling~\cite{ref2,ref45}. LSTM-based approaches typically use a rolling LSTM to encode observed video sequences, but they struggle with capturing long-horizon temporal dependencies, despite enhancements like goal-based learning and diverse attention mechanisms. Transformer-based methods~\cite{ref18,ref47,ref49}, which employ global attention mechanisms, have gained traction for their ability to operate in both uni-modal and multi-modal settings, incorporating RGB, optical flow, and object-based features. Girdhar et al.~\cite{ref43} proposed anticipative video transformers with a  self-attention design. HRO~\cite{ref41} leverage novel caching mechanisms to store long-term prototypical activity semantics. However, memory bank methods, while effective, are computationally expensive and require significant memory and processing power, making them less efficient for real-time applications. Also method lack  unified attention block may not fully exploit the unique properties between different modalities.

 \subsection{Worker Assistance Systems}
\label{WASystem}
Assistance systems in manufacturing are designed to support workers by enhancing their capabilities without replacing or overriding them~\cite{mark2021worker}. These systems aim to address deficits such as age-related limitations, skill gaps, or disabilities, ultimately improving productivity by reducing errors and streamlining workflows. The primary objective is to provide context-aware, easily accessible information that aids in task execution while minimizing cognitive load and preventing potential anomalies. Mura et al.~\cite{dalle2016integrated} introduced a manual assembly workstation that detects errors in component selection and orientation, providing immediate corrective instructions. Faccio et al.\cite{faccio2019real} proposed a system that visually guides workers by monitoring tool and component usage during assembly. Wang et al.~\cite{wang2021smart} introduced a deep learning-based angle-monitoring system that provides real-time feedback on tool angles, reducing human error on the assembly line. However, many existing monitoring systems rely on rule-based approaches, limiting their effectiveness in complex assembly lines. In dynamic workflows, anticipatory capabilities using graph-based insights are essential for improving generalization and task efficiency. By proactively guiding operators, an effective system enhances decision-making and ensures smooth execution in complex assembly tasks.

\section{Proposed Approach}
\label{PA}

\begin{figure}[!]
\centering
\includegraphics[width=0.48\textwidth]{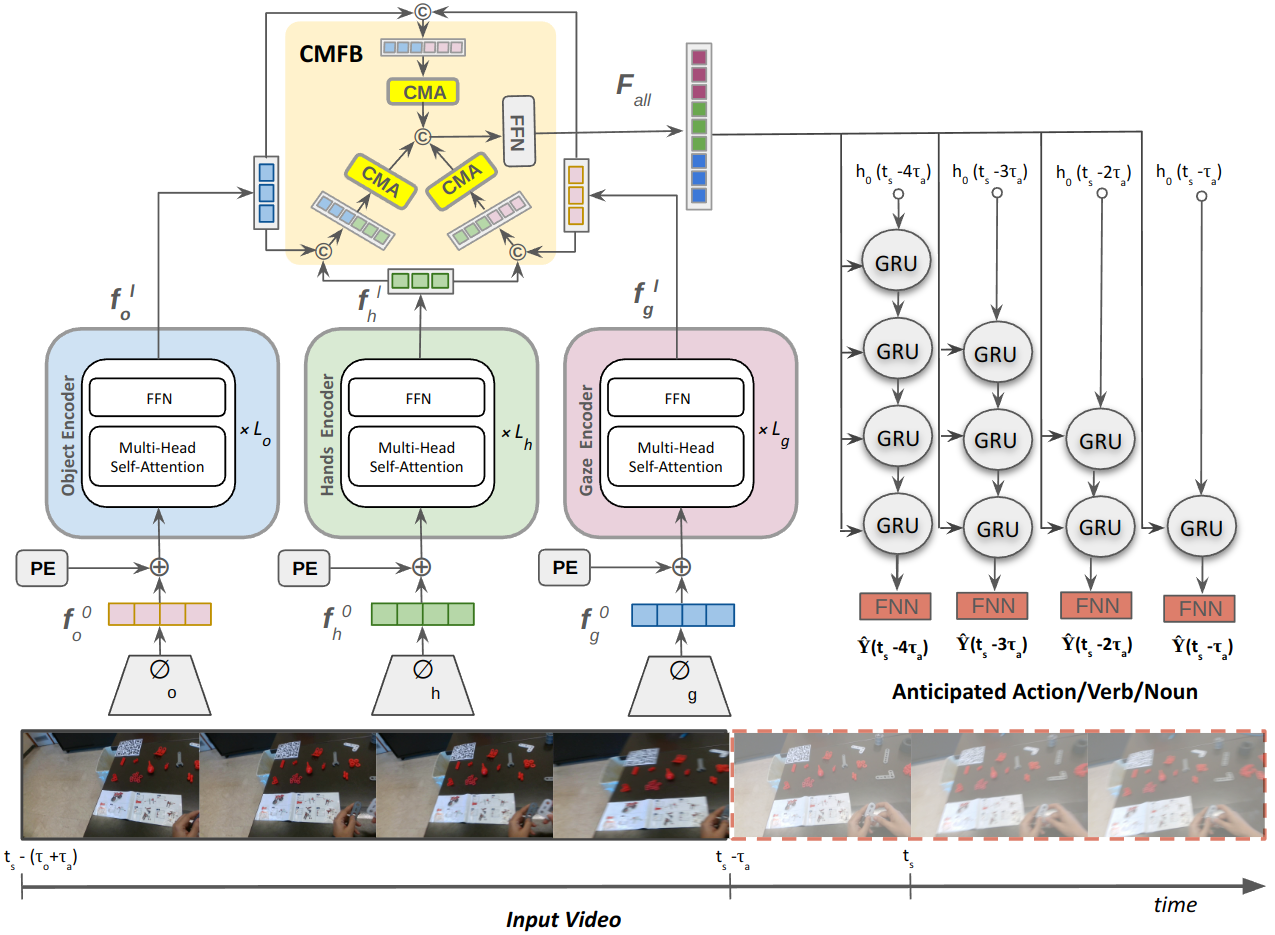}
\caption{The architecture of the proposed MMTF-RU framework. Input video features are extracted via a TSN~\cite{ref35}, resulting modality-specific features ($\bm{f}{o}^{0}$, $\bm{f}{h}^{0}$, $\bm{f}{g}^{0}$). These, along with positional embeddings (PE), are processed by transformer encoders to produce transformed features ($\bm{f}{o}^{l}$, $\bm{f}{h}^{l}$, $\bm{f}{g}^{l}$). The CMFB integrates features across modalities, and GRUs generate temporal decoder features based on anticipation time $\tau_a$. Finally, these features are classified to predict the next action, verb, or noun ($\hat{Y}$).}

\label{arch}
\end{figure}

Anticipating actions involves the prediction of future activities based on visual information extracted from current and preceding frames. Our proposed MMTF-RU model, illustrated in Fig.~\ref{arch}, comprises three primary components: transformer-based encoder, CMFB for integrating inputs from different modalities, and GRU-based decoder for generating future action predictions.  Fig.~\ref{overview} illustrates the problem setup,  given  a video sequence $\bm{V}$ with $T$ frames, the initial $[T - \frac{t_{s} - \tau_{a}}{\tau}]$ frames are observed and used during the encoding stage, while the remaining $\frac{t_{s} - \tau_{a}}{\tau}$ frames are reserved for anticipating subsequent actions. Here,   $\tau$ represents the time-step between each consecutive pair of frames during the anticipation period $\tau_{a}$ . During the encoding stage, the model condenses the semantic content of input video snippets without making predictions. In the decoder stage, using a recurrent network, the model continues processing semantic information and generates action anticipation predictions at various times $\tau_{a} \in \{2s, 1.75s, 1.5s, 1.25s, 1s, 0.75s, 0.5s, 0.25s\}$, following~\cite{ref1,ref2}. The predictions generated by the MMTF-RU model  at $\tau_{a} =1s$  are then utilized by the OAMU, which integrates a graph-based method for anomaly prevention and next action guidance. A reference graph built on Markov principles generates subgraphs to detect anomalies by comparing predicted and observed action transitions. This enables real-time guidance by suggesting corrective actions when deviations occur. The TWSA metric is used to assess the efficiency of these guided actions, ensuring the model not only predicts actions accurately but also enhances performance in real-world scenarios.

\subsection{MMTF-RU} 
\subsubsection{Encoding} 
Given the input video sequence $\bm{V} \in \mathbb{R}^{T \times C \times H \times W}$ to be processed to  high  level  modality specific representations $\bm{f}_{m}^{0} \in \mathbb{R}^{T \times D}$  for   $T$ frames obtained through a Temporal Segment Network (TSN)~\cite{ref35} $\phi_{m}(.)$, where $m \in \{o,h,g\}$  represents the considered modalities, namely object-based features, hands-based features, and gaze features. TSN is employed to capture long-range temporal dependencies and enhance the robustness of the extracted features by effectively handling temporal information.

Subsequently passed to a transformer block consisting of $L_{o}$, $L_{h}$, and $L_{g}$ layers for each modalities. Each encoder layer is comprised of a Multi-Head Self-Attention (MSA), as defined in Eq.~\ref{eq3}, followed by layer normalization (LN) and a feed-forward network (FFN) with a residual connection~\cite{ref3}. We compute the dot products attention with each input modality representation $\bm{f}_{m}^{0}$ defined as follows:  

\begin{equation}\label{eq1}
\text{E}_{\text{Attn}}(\textbf{X}) = \sigma\left(\frac{(\textbf{X}\mathbf{W}_{i}^{Q})(\textbf{X}\mathbf{W}_{i}^{K})^T}{\sqrt{d_k}}\right) \textbf{X}\mathbf{W}_{i}^{V}
\end{equation}

\noindent Where \textbf{X} denotes the intermediate variable, which is defined as \(\textbf{X} = \bm{f}_{m}^{0} + \bm{P}\) as stated later in Eq.~\ref{eq4}. The attention mechanism relies on a trainable associative memory with queries, keys, and values pairs defined as linear layers $\mathbf{W}_{i}^{Q}, \mathbf{W}_{i}^{K}, \mathbf{W}_{i}^{V}$ applied on the input sequence at the $i^{th}$ head. Here, $\mathbf{W}_{i}^{Q}, \mathbf{W}_{i}^{K}, \mathbf{W}_{i}^{V} \in \mathbb{R}^{D \times \frac{D}{N}}$, where $N$ is the number of heads, and $\sigma$ denotes the softmax function. The term $\frac{1}{\sqrt{d_k}}$ serves as a scaling factor to enhance training stability and accelerate convergence.

\begin{equation}\label{eq3} 
\text{MSA}(\textbf{X}) = \left[\text{E}_{\text{Attn}}(\textbf{X})_1, \text{E}_{\text{Attn}}(\textbf{X})_2, ..., \text{E}_{\text{Attn}}(\textbf{X})_N\right]\mathbf{W}_o
\end{equation}

 \noindent  where  $\mathbf{W}_o \in \mathbb{R}^{D \times D}$ represents the output linear projection layer. 
 

Following this, a feed-forward network (FFN) with Leaky-ReLU activation is employed. Simultaneously, layer normalization and residual connections are applied. To retain temporal information, a learnable positional embedding $\bm{P}$ is utilized in conjunction with the modality input $\bm{f}_{m}^{0}$. The final output token $\bm{f}_{m}^{l}$ from the $l^{th}$ encoder layer can be expressed as:

\begin{equation}\label{eq4}
\bm{f}_{m}^{l} = \text{FFN}\left(\text{Norm}\left(\text{MSA}(\bm{f}_{m}^{0} + \bm{P})\right) + \bm{f}_{m}^{0}\right)
\end{equation}

\subsubsection{CMFB} 
The proposed method utilizes the intermediate modality features extracted from the final layer of the transformer encoder. The modality features extracted from the transformer encoder are combined in pairs as pairwise modality features, denoted as $\bm{F}_{pw}^{k}$ and defined in Eq.~\ref{eq5}. This formulation involves a total of $k = 3$ unique pairs across the three modalities. These refined features are passed through an MSA block and subsequent functions to discern the correlation between cross-modalities $\bm{F}_{cma}^{k}$ defined in Eq.~\ref{eq6}. The features are concatenated and undergo a transformation using the function $\text{L}_{\text{proj}}(\cdot) = \text{LeakyReLU} \leftarrow \text{BN} \leftarrow \text{Conv}_{1 \times 1}$, resulting in the formation of $\bm{F}_{all}$ as specified in Eq.~\ref{eq7}, thereby deriving the final fused features. The CMFB  applies MSA to capture interactions between modalities, improving feature representation by pairing modality features to explore complementary aspects. This fusion simplifies interaction modeling and offers a flexible, scalable approach to understanding complex relationships. Early fusion within CMFB allows the model to learn joint representations from the beginning, enhancing the overall interaction understanding.

\begin{equation}\label{eq5}
\bm{F}_{pw}^{k} = \left\{ \text{Concat}\left(\bm{f}_{i}^{l}, \bm{f}_{j}^{l}\right) \mid i \neq j \right\}
\end{equation}

\begin{equation}\label{eq6}
    \bm{F}_{cma}^{k} = \left[ \text{Norm}(\text{MSA}(\bm{x})) + \bm{x} \, \big\vert \, \bm{x} \in \bm{F}_{pw}^{k} \right]
\end{equation}

\begin{equation}\label{eq7}
    \bm{F}_{\text{all}} = \text{L}_{\text{proj}}\left(\text{Concat}\left(\bigcup_{i=1}^{k} \bm{F}_{cma}^{i}\right)\right)
\end{equation}

\subsubsection{Decoding} 
 We use  GRU~\cite{ref4}  for decoding due to its flexibility in handling variable anticipation times. As a recurrent neural network, GRU can adeptly adapt to different temporal dynamics within sequential data. Similar to~\cite{ref1}, the anticipation stage involves iterating over the hidden vectors of the GRU at the current time-step and processing the representation of the current video snippet $\bm{F}_{\text{all}}$. This iteration occurs for a specific number of times, denoted by $s = \frac{t_{s} - \tau_{a}}{\tau}$, which corresponds to the number of time-steps needed to reach the beginning of the action. The initialization of the hidden layer is depicted in Eq.~\ref{eq2}, which takes the concatenated modality features followed by the $\text{L}_{\text{hid}}(\cdot)$ transformation function, akin to $\text{L}_{\text{proj}}(\cdot)$. The output from the last hidden layer of the decoder, as described in Eq.~\ref{eq8}, serves as the input to generate logits for future actions, denoted as $\bm{\hat{Y}} \in \mathbb{R}^{t \times \text{class}}$. This is accomplished using a fully connected layer with weights $\mathbf{W}_A$, as outlined in Equation \ref{eq9}.

\begin{equation}\label{eq2}
    \mathbf{h}_0  = \text{L}_{\text{hid}}\left(\text{Concat}\left(\bm{f}_{o}^{0}, \bm{f}_{h}^{0}, \bm{f}_{g}^{0}\right)\right)
\end{equation}

\begin{equation}\label{eq8}
\mathbf{h}_{i+1} = \text{GRU}(\bm{F}_{all}, \mathbf{h}_i) \quad \text{for}\, i = 0, 1, \ldots, s
\end{equation}

\begin{equation}\label{eq9}
\bm{\hat{Y}} = \mathbf{W}_A \cdot \mathbf{h}_{s}
\end{equation}

\subsection{OAMU}

\begin{figure}[!h]
\centering
\subfloat[]{\includegraphics[width=0.49\textwidth]{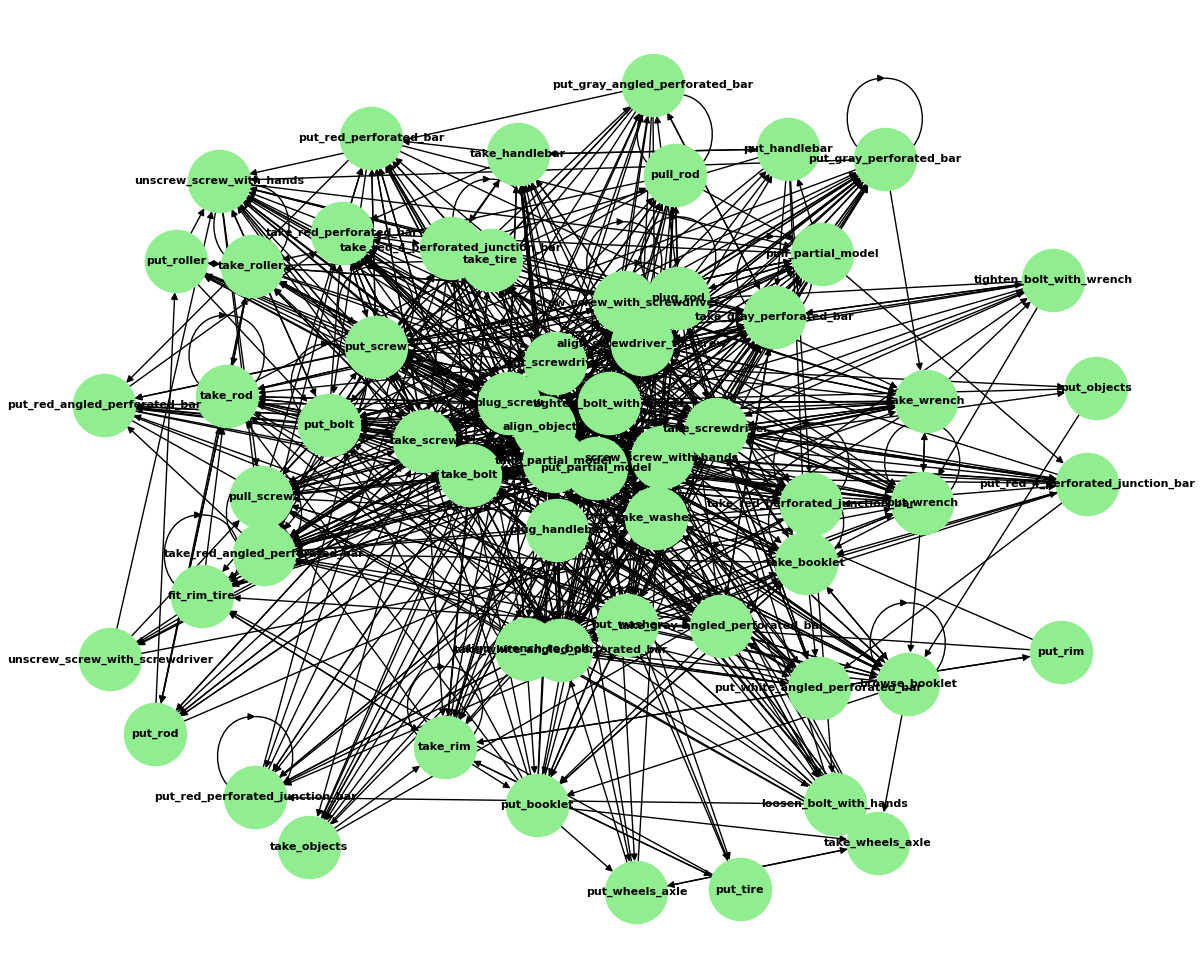}\label{trans_imgs:sub1}} \\
\subfloat[]{\includegraphics[width=0.49\textwidth]{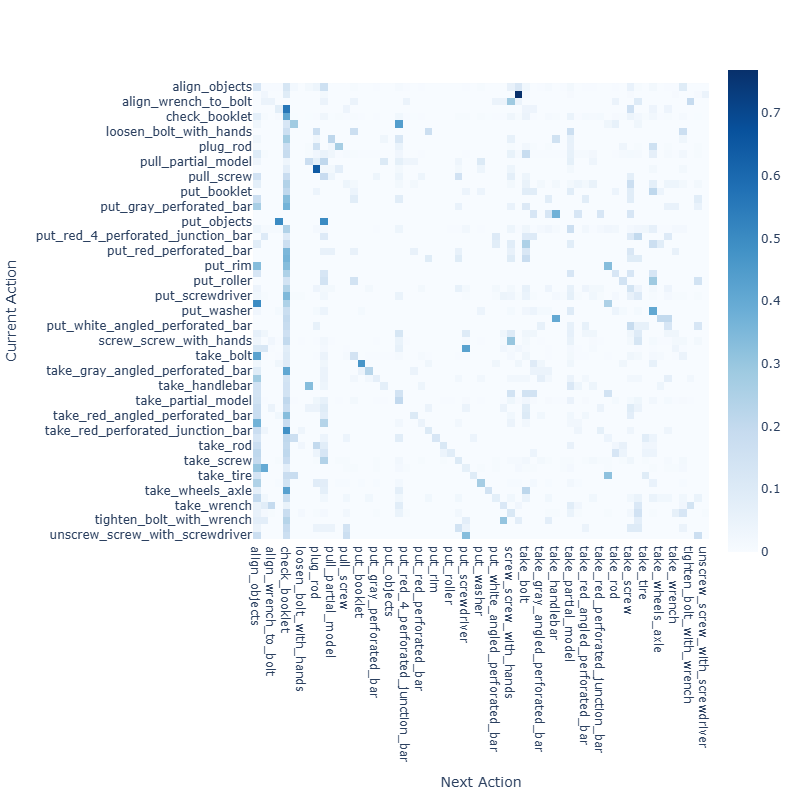}\label{trans_imgs:sub2}}  

\caption{ (a) Reference graph and (b) transition heatmaps in the Meccano dataset.} \label{trans_imgs}
\end{figure}

Our approach models action sequences as a first-order Markov chain, as outlined in Algo. \ref{alg:construct_reference_graph}. This is represented by a directed graph \( G = (V, E, w) \), where each node corresponds to an action and each edge \( (u, v) \) indicates a transition between actions. The graph, depicted in  Fig.~\ref{trans_imgs:sub1}, includes distinct representations for actions of the Meccano~\cite{ref5} dataset. The weights \( w(u, v) \) reflect normalized transition frequencies, calculated as the ratio of observed transitions \( (s_k, s_{k+1}) \) in the sequence \( \mathbf{S} = [s_1, s_2, \ldots, s_n] \) to the total transitions \( T \), as shown in  Fig.~\ref{trans_imgs}(b). This normalization produces  transition probability \( P(s_{k+1} | s_k) \), upholding  Markov property and forming the basis for the subsequent algorithms in our framework. \textbf{Note:} ``Action'' refers to a combination of verb and noun(s), and may be replaced by ``noun'' or ``verb'' depending on the context.

\begin{algorithm}[!h]
\caption{Constructing the reference graph from a action sequences.}
\label{alg:construct_reference_graph}
\scriptsize
\begin{algorithmic}
\Statex \textbf{Input:} Sequence of actions \( \mathbf{S} = [s_1, s_2, \ldots, s_n] \)
\Statex \textbf{Output:} Directed graph \( G = (V, E, w) \) with normalized transition weights

\Statex \textbf{Initialization:}
\Statex \hspace{1em} Initialize an empty directed graph \( G = (V, E) \)
\Statex \hspace{1em} \( T \gets 0 \)  \Comment{Total number of transitions}

\For{\( k = 1 \) to \( n - 1 \)}    \Comment{Iterate over action pairs}
    \State \( (s_k, s_{k+1}) \gets (\mathbf{S}[k], \mathbf{S}[k+1]) \)
    \If{\( (s_k, s_{k+1}) \in E \)}
        \State \( w(s_k, s_{k+1}) \gets w(s_k, s_{k+1}) + 1 \)
    \Else
        \State Add edge \( (s_k, s_{k+1}) \) to \( G \) with \( w(s_k, s_{k+1}) \gets 1 \)
    \EndIf
    \State \( T \gets T + 1 \)
\EndFor

\For{each edge \( (u, v) \in E \)}
    \State \( w(u, v) \gets \frac{w(u, v)}{T} \)  \Comment{Normalized edge weights}
\EndFor

\Statex \textbf{Return:} Graph \( G \)
\end{algorithmic}
\end{algorithm}

\begin{algorithm}[!h]
\caption{Proactive guidance using MMTF-RU top-5 predictions and reference graph, supported by an action dictionary.}
\label{alg:guidance_operator_subgraph}
\scriptsize
\begin{algorithmic}
\Statex \textbf{Input:} Reference graph \( G = (V, E, w) \), Initial state \( t_0 \), Action dictionary \( D \)
\Statex \textbf{Output:} Subgraph \( G_{\text{sub}} = (V_{\text{sub}}, E_{\text{sub}}, w_{\text{sub}}) \), Next action recommendation \( a_{\text{next}} \), and  operator feedback.

\Statex \textbf{Initialization:}
\Statex \hspace{1em} Initialize empty subgraph \( G_{\text{sub}} = (V_{\text{sub}}, E_{\text{sub}}) \)
\Statex \hspace{1em} Initialize empty set of valid graph-based actions \( G_{\text{valid}} \leftarrow \emptyset \)
\Statex \hspace{1em} Initialize empty set of valid model-based actions \( P_{\text{valid}} \leftarrow \emptyset \)


\State \( \text{Succ}(t_0) \gets \{v \mid (t_0, v) \in E\} \)  \Comment{Set of successors of \( t_0 \) in \( G \)}
\For{each \( s_j \in \text{Succ}(t_0) \)}
    \State Add edge \( (t_0, s_j) \) to \( G_{\text{sub}} \) with weight \( w_{\text{sub}}(t_0, s_j) \leftarrow w(t_0, s_j) \)
\EndFor
 
\State \( p_j \gets \frac{w_{\text{sub}}(t_0, s_j)}{ \sum_{s_j \in \text{Succ}(t_0)} w_{\text{sub}}(t_0, s_j)} \)  \Comment{Probability of transition to successor \( s_j \)}
    
\State \( \text{Succ}_{\text{sorted}}(t_0) \gets \left\{(s_j, p_j) \mid s_j \in \text{Succ}(t_0)\right\}_{\downarrow p_j} \)
\State \( G_{\text{sub}} \gets G_{\text{sub}} \cup \ \text{Succ}_{\text{sorted}}(t_0) \)
\State  \( Y(t_0) \gets \text{MMTF-RU}(t_0) \) \Comment{Anticipated Top-5 actions}

\State \( G_{\text{valid}}(t_0) \gets G_{\text{sub}}(t_0) \cap D \)
\State \( Y_{\text{valid}}(t_0) \gets  Y(t_0)  \cap D \)


\State \( A_{\cap} \gets G_{\text{valid}}(t_0) \cap P_{\text{valid}}(t_0) \)

\If{\( A_{\cap} \neq \emptyset \)}
    \State \( a_{\text{next}} \gets \arg \min_{a \in A_{\cap}} \left( \text{rank}_{G}(a) + \text{rank}_{P}(a) \right) \)
    \State \textbf{Return:} \( a_{\text{next}} \) \Comment{Operator next action}
\Else
    \State \textbf{Feedback:} Ask the operator to repeat the previous action, and suggest next actions \( G_{\text{valid}}(t_0) \).
\EndIf

\State Update \( G_{\text{sub}} \) and repeat for the next observation \( t_{\text{next}} \).

\end{algorithmic}
\end{algorithm}

\begin{algorithm}[!h]
\caption{Proactive guidance and anomaly score prediction using MMTF-RU top-1 predictions and reference graph, supported by an entropy-informed confidence mechanism.}
\label{alg:subgraph_and_anomaly}
\scriptsize
\begin{algorithmic}
\Statex \textbf{Input:} Reference graph \( G = (V, E, w) \), Test sequence \( \mathbf{t} = [t_1, t_2, \ldots, t_m] \),  Maximum number of top next actions \( k =  5\)
\Statex \textbf{Output:} Subgraph \( G_{\text{sub}} = (V_{\text{sub}}, E_{\text{sub}}, w_{\text{sub}}) \), Anomaly scores \( \mathbf{\mathcal{A}} \), Top-5 next actions for each state \( \mathbf{\mathcal{N}} \)

\Statex \textbf{Initialization:}
\Statex \hspace{1em} Initialize empty directed graph \( G_{\text{sub}} = (V_{\text{sub}}, E_{\text{sub}}) \)
\Statex \hspace{1em} Initialize empty list \( \mathbf{\mathcal{A}} \leftarrow [] \)  \Comment{List to store non-zero anomaly scores}
\Statex \hspace{1em} Initialize empty list \( \mathbf{\mathcal{N}} \leftarrow [] \)  \Comment{List to store corresponding Top-5 next actions}

\For{\( i = 1 \) to \( m-1 \)}
    \State \( t_i \gets \mathbf{t}[i] \)  \Comment{Current action}
    \State \( t_{i+1} \gets \mathbf{t}[i+1] \)  \Comment{Next action}
    \State \( \text{Succ}(t_i) \gets \{v \mid (t_i, v) \in E\} \)  \Comment{Set of successors of \( t_i \) in \( G \)}

    \For{each \( s_j \in \text{Succ}(t_i) \)}
        \State Add edge \( (t_i, s_j) \) to \( G_{\text{sub}} \) with weight \( w_{\text{sub}}(t_i, s_j) \leftarrow w(t_i, s_j) \)
    \EndFor

    \State \( p_j \gets \frac{w_{\text{sub}}(t_i, s_j)}{ \sum_{s_j \in \text{Succ}(t_i)} w_{\text{sub}}(t_i, s_j)} \)  \Comment{Probability of transition to successor \( s_j \)}
    \State \( H(t_i) \gets -\sum_{s_j \in \text{Succ}(t_i)} p_j \log(p_j) \)  \Comment{Entropy of the successor action's}


    \State \( \text{Succ}_{\text{sorted}}(t_i) \gets \left\{(s_j, p_j) \mid s_j \in \text{Succ}(t_i)\right\}_{\downarrow p_j} \)
    \State \( N(t_i) \gets \{ (s_j, p_j) \}_{j=1}^{k}, \quad (s_j, p_j) \in \text{Succ}_{\text{sorted}}(t_i) \)

    \State \( r(t_{i+1})  \gets \text{rank}(t_{i+1} \mid \text{Succ}_{\text{sorted}}(t_i)) + 1 \)
    \State \( p(t_{i+1}) \gets \frac{w_{\text{sub}}(t_i, t_{i+1})}{\sum_{s_j \in \text{Succ}(t_i)} w_{\text{sub}}(t_i, s_j)} \)  \Comment{Probability of observed transition}
    \State \(  c(t_{i+1}) \gets  1 - \frac{-p(t_{i+1}) \log(p(t_{i+1}))}{H(t_i)} \)  \Comment{Observed action certainty}
    \State  \(  a(t_{i+1}) \gets    \left( \frac{\log(\text{r}(t_{i+1}))}{\log( |\text{Succ}(t_i)|))} \right)  \times   
      \left( 1 - \frac{-p(t_{i+1})}{\max\limits_{s_j \in \text{Succ}(t_i)} p_j}  \right) \times    c(t_{i+1}) \) 

    \If{\( a(t_{i+1}) > 0 \)}  
        \State \( \mathbf{\mathcal{A}} \gets \mathbf{\mathcal{A}} \cup \{a(t_{i+1})\} \) \Comment{Append anomaly score to the list}
        \State \( \mathbf{\mathcal{N}} \gets \mathbf{\mathcal{N}} \cup \{N(t_i)\} \) \Comment{Append Top-5 actions to the list}
    \EndIf
\EndFor

\Statex \textbf{Return:} Subgraph \( G_{\text{sub}} \), Anomaly scores \( \mathbf{\mathcal{A}} \), Top-5 next actions \( \mathbf{\mathcal{N}} \)
\end{algorithmic}
\end{algorithm}

The framework integrates graph-based recommendations with an anticipation model to guide operator next actions and prevent anomalies (Algo.~\ref{alg:guidance_operator_subgraph}). Beginning with the current state \( t_0 \), a subgraph \( G_{\text{sub}} \) is extracted from the reference graph \( G \), identifying potential next actions based on transition probabilities. Concurrently, the MMTF-RU anticipation model provides Top-5 predictions \( Y(t_0) \) to enhance decision-making. Valid actions from both the graph and model are filtered through the action dictionary \( D \), and the set of common actions \( A_{\cap} \) is obtained by minimizing a combined ranking from both sources. This method harnesses the flexibility of Top-5 predictions while ensuring robust guidance. In cases where \( A_{\cap} = \emptyset \), the operator is prompted to repeat the previous action, ensuring process continuity. The system continuously updates as new observations \( t_{\text{next}} \) are made, improving adaptability throughout the task sequence.

Next, we introduce a next action and anomaly prevention framework outlined in Algo.~\ref{alg:subgraph_and_anomaly}, where the framework generates an anomaly severity score and processes a test sequence \( \mathbf{t} = [t_1, t_2, \ldots, t_m] \) generated by the MMTF-RU model. A subgraph \( G_{\text{sub}} \) is extracted from the reference graph \( G \), focusing on the observed transitions, which result in the prediction of probable subsequent actions \( \text{Succ}(t_i) \). The operator's intended action, as predicted by MMTF-RU, is checked against these successors, and guidance is provided for the next action. The anomaly score \( a(t_{i+1}) \) for the operator's anticipated action is computed using three factors: the rank \( r(t_{i+1}) \) of the observed action among the Top-1 successor, the transition probability \( \left( 1 - \frac{-p(t_{i+1})}{\max_{s_j \in \text{Succ}(t_i)} p_j} \right) \), and the certainty \( c(t_{i+1}) \) derived from entropy. This formulation, combining rank, probability deviation, and certainty, offers a robust metric for detecting deviations from expected patterns.

In dynamic industrial settings, it is essential to identify the Top-5 likely next actions while ensuring efficient execution. The median reference time \( t_{\text{reference}}(a_i) \), evaluated from the training set, provides a robust benchmark. We define  Time-Weighted Sequence Accuracy (TWSA) as:
 
\begin{equation}\label{eq10}
\text{TWSA}_i = \min\left(\frac{t_{\text{reference}}(a_i)}{t_{\text{actual}}(a_i)}, 1\right) \times 1\left(\text{Seq}_i = \text{Seq}_{\text{optimal}}\right)
\end{equation}

Here, \( \text{Seq}_i \) is the actual sequence of actions performed, and \( \text{Seq}_{\text{optimal}} \) is the expected sequence. This formula ensures that TWSA reflects both the correctness within the Top-5 predicted actions and adherence to optimal execution time, which is critical for maintaining efficiency in time-sensitive workflows.

\section{Experiments and Results}
\label{CD}

\subsection{Datasets} 
The \textbf{Meccano}~\cite{ref5} dataset is a multi-modal egocentric dataset collected in an industrial-like setting to analyze human-object interactions during instructional tasks. It combines gaze, object-centric, and hand-centric features, providing a rich exploration of human activities in industrial environments. The dataset includes 20 object classes (16 toy component classes and 2 tool classes: screwdriver and wrench), 12 verb classes, and 61 action classes. It consists of 20 videos, with 11 used for training and 9 for validation and testing.  

The \textbf{EPIC-Kitchens-55}~\cite{ref6} dataset contains 55 hours of video recordings of daily kitchen activities from 32 participants. It features 125 verb classes and 352 noun classes, forming 2,513 unique action labels from (verb, noun) pairs. Following the setup in ~\cite{ref7}, we divide the 28,472 activity segments into 23,493 for training and 4,979 for validation.

\subsection{Implementation Details}

Our model follows the setup in ~\cite{ref7}, with 14 total time-steps ($T$) before the initiation of the next action $Y$ at $t_{s}$. Each time-step occurs at intervals of $\tau = 0.25s$, with the observed steps lasting 6 units ($\tau_{o}$). Anticipation is performed for the next 8 time-steps ($\tau_a$) at intervals of 2s, 1.75s, 1.5s, down to 0.25s. The model is trained for 100 epochs with a batch size of 128, using SGD optimization, an initial learning rate of 0.001, and momentum of 0.9. 

\subsection{Comparison with State-of-the-Art Methods}

\setlength{\intextsep}{10pt}
\begin{table}[h]
\centering
\caption{Comparison results for the action anticipation task on the Meccano~\cite{ref5} dataset are presented in terms of Top-1 and Top-5 accuracy (in \%), considering various anticipation time intervals $\tau_{a}$. The best results are highlighted in boldface.}\label{tab1}
\resizebox{0.49\textwidth}{!}{%
\begin{tabular}{cccccccccc}
\hline
\textbf{Method} & \textbf{2s} & \textbf{1.75s} & \textbf{1.50s} & \textbf{1.25s} & \textbf{1s} & \textbf{0.75s} & \textbf{0.50s} & \textbf{0.25s} \\ \hline
\multicolumn{9}{c}{\textbf{Top-1 Accuracy}} \\ \hline
RU-LSTM~\cite{ref7}              & 23.37       & 23.48          & 23.30          & 23.97          & 24.08      & 24.50          & 25.60          & 28.87          \\
VLMAH~\cite{ref2}                & 24.75       & 24.35          & 24.22          & 22.79          & 28.90      & 25.29          & 26.47          & 29.12          \\
MMTF-RU (proposed)               & \textbf{27.80} & \textbf{28.05} & \textbf{28.83} & \textbf{29.22} & \textbf{29.75} & \textbf{30.18} & \textbf{30.50} & \textbf{30.50} \\ \hline
Improvement                      & +3.05       & +3.70          & +4.61          & +5.25          & +0.85      & +4.89          & +4.03          & +1.38          \\ \hline
\multicolumn{9}{c}{\textbf{Top-5 Accuracy}} \\ \hline
RU-LSTM~\cite{ref7}              & 54.65       & 55.99          & 56.56          & 57.73          & 58.23      & 59.96          & 61.31          & 63.40          \\
VLMAH~\cite{ref2}                & 54.23       & 55.16          & 53.09          & 53.98          & 58.13      & 53.16          & 56.71          & 58.01          \\
DCR~\cite{ref45}                 & $-$         & $-$            & $-$            & $-$            & 56.7       & $-$            & $-$            & $-$            \\
Ub-DCR~\cite{ref44}              & $-$         & $-$            & $-$            & $-$            & 60.3       & $-$            & $-$            & $-$            \\
Ub-RULSTM~\cite{ref44}           & 60.30       & 61.50          & 61.20          & 62.30          & 62.70      & 63.90          & 64.00          & \textbf{65.70}          \\
MMTF-RU (proposed)               & \textbf{63.83} & \textbf{64.11} & \textbf{64.47} & \textbf{65.18} & \textbf{64.46} & \textbf{65.78} & \textbf{65.82} & 64.93 \\ \hline
Improvement                      & +3.53       & +2.61          & +3.27          & +2.88          & +1.76      & +1.88          & +1.82          & -0.77          \\ \hline
\end{tabular}
}
\end{table}
 
\begin{figure}[htbp]
\centering
\subfloat[]{\includegraphics[width=0.45\textwidth]{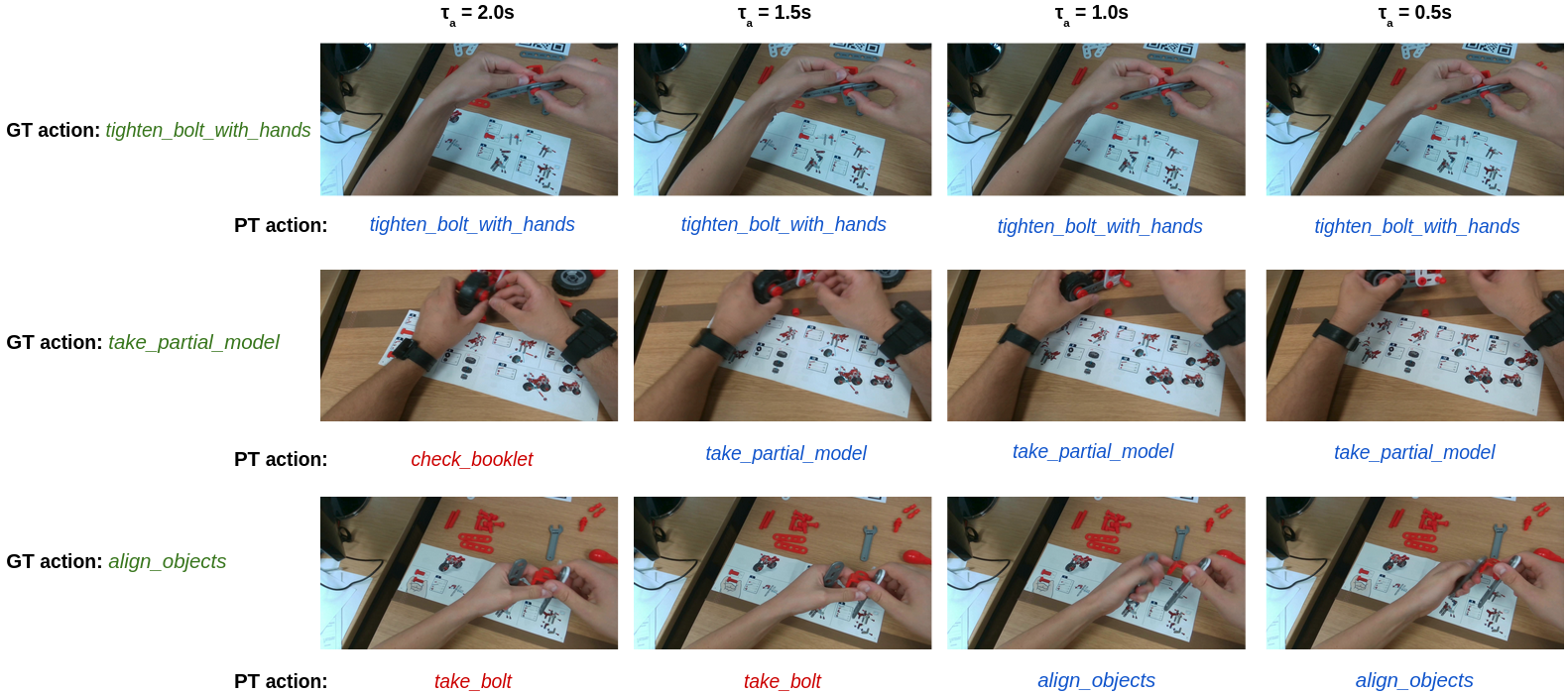}\label{fig:sub1}}
\hfill
\subfloat[]{\includegraphics[width=0.45\textwidth]{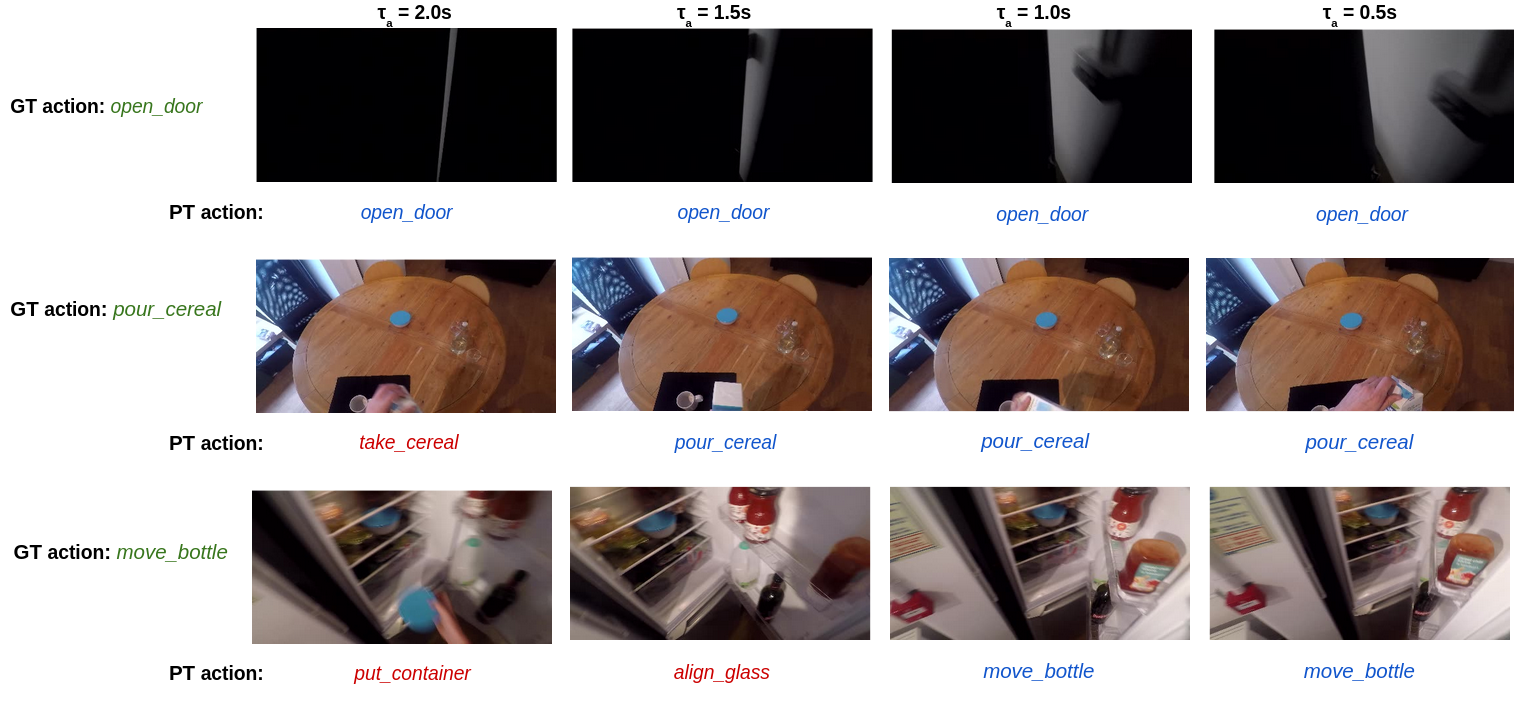}\label{fig:sub2}}

\caption{Visualization of Top-1 action anticipation results for (a) Meccano and (b) EPIC-Kitchens-55 datasets. Ground truth (GT) actions are highlighted in blue, correct predictions (PT) in green, and incorrect predictions in red.}

\label{fig:test}
\end{figure}

\noindent \textbf{Results on Meccano.} Table~\ref{tab1} compares the proposed MMTF-RU model on the Meccano dataset with SOTA methods. We evaluated Top-1 and Top-5 activity accuracy for the next segment across 8 anticipation times $\tau_a$ ranging from 0.25s to 2s, utilizing gaze, object-centric, and hand-centric features as identified in ~\cite{ref5}. Our model, combining all modalities, outperforms baseline methods, including RU-LSTM~\cite{ref5}, DCR~\cite{ref45}, Ub-DCR~\cite{ref44}, Ub-RULSTM~\cite{ref44}, and VLMAH~\cite{ref2}, across all anticipation times. At $\tau_a = 1s$, MMTF-RU achieves 29.75\% Top-1 and 64.46\% Top-5 accuracy, improving by +0.85\% and +1.76\% over VLMAH~\cite{ref2} and Ub-RULSTM~\cite{ref44}. Fig.~\ref{fig:sub1} visualizes Top-1 results for $\tau_a = 0.5s, 1s, 1.5s, 2s$, showing both accurate and incorrect predictions. Model failures often result from scene occlusions, such as hands covering target objects or actions being out of frame. Table~\ref{tab2} shows Top-1 and Top-5 accuracy for verbs and nouns at $\tau_a = 1s$, with our approach significantly outperforming SOTA in both categories, except for a slight reduction in Top-5 verb accuracy.

\begin{table}[h]
\centering
\caption{Comparison results for the noun and verb anticipation tasks on the Meccano~\cite{ref5} dataset are presented in terms of Top-1 and Top-5 accuracy (in \%) at an anticipation time of $\tau_a = 1s$. The best  results are highlighted in boldface.}\label{tab2} 
\resizebox{0.36\textwidth}{!}{%
\begin{tabular}{lllll}
                                 &                 &                 &                 &                 \\ \hline
\multirow{2}{*}{\textbf{Method}} & \multicolumn{2}{c}{\textbf{Noun}} & \multicolumn{2}{c}{\textbf{Verb}} \\ \cline{2-5} 
                                 & \textbf{Top-1}  & \textbf{Top-5}  & \textbf{Top-1}  & \textbf{Top-5}  \\ \hline
RU-LSTM~\cite{ref7}                          & 30.53           & 66.49           & 35.92           & \textbf{93.19}  \\
VLMAH~\cite{ref2}                            & 31.25           & 76.30           & 35.96           & 92.88           \\
MMTF-RU (proposed)                              & \textbf{37.77}  & \textbf{85.82}  & \textbf{39.89}  & 92.09           \\ \hline
Improvement                      & +6.52           & +9.52           & +3.93           & -1.1           \\ \hline
\end{tabular}
}
\end{table}

\begin{table}[]
\centering
\caption{Comparison of action anticipation results on the EPIC-Kitchens-55~\cite{ref6} validation set. The best and second-best performances are highlighted in bold and underlined, respectively.}
\label{tab3}
\resizebox{0.50\textwidth}{!}{%
\begin{tabular}{ccccccccccc}
\hline
\multicolumn{1}{c}{\multirow{2}{*}{\textbf{Method}}} & \multicolumn{4}{c}{\textbf{Top-5 Accuracy}} & \multicolumn{6}{c}{\textbf{Top-5 Accuracy @1s}} \\ \cline{2-11} 
\multicolumn{1}{c}{} & \textbf{2s} & \textbf{1.5s} & \textbf{1s} & \textbf{0.5s} & \multicolumn{3}{c}{\textbf{Acc. \% @ 1s}} & \multicolumn{3}{c}{\textbf{M. Rec. \% @ 1s}} \\ \cline{6-11} 
\multicolumn{1}{c}{} & & & & & \textbf{Verb} & \textbf{Noun} & \textbf{Act} & \textbf{Verb} & \textbf{Noun} & \textbf{Act} \\ \hline
RL {~\cite{ref9}} & 25.95 & 27.15 & 29.61 & 31.86 & 76.80 & 44.50 & 29.60 & 40.80 & 40.90 & 10.60 \\
EL {~\cite{ref10}} & 24.68 & 26.41 & 28.56 & 31.50 & 75.70 & 43.70 & 28.60 & 38.70 & 40.30 & 8.60 \\
RU-LSTM {~\cite{ref7}} & 29.44 & 32.24 & 35.32 & 37.37 & 79.60 & 51.80 & 35.30 & 43.80 & 49.90 & 15.10 \\
SRL {~\cite{ref12}} & 30.15 & 32.36 & 35.52 & 38.60 & $-$ & $-$ & 35.50 & $-$ & $-$ & $-$ \\
LAI {~\cite{ref13}} & $-$ & 32.50 & 35.60 & 38.50 & $-$ & $-$ & 35.60 & $-$ & $-$ & $-$ \\
TempAgg {~\cite{ref11}} & 30.90 & 33.70 & 36.40 & 39.50 & $-$ & $-$ & 35.60 & $-$ & $-$ & $-$ \\
AVT {~\cite{ref43}} & $-$ & $-$ & $-$ & $-$ & 79.90 & 54.00 & 37.60 & $-$ & $-$ & $-$ \\
Ub-RULSTM {~\cite{ref44}} & 30.10 & 33.10 & 35.80 & 38.40 & \underline{80.40} & 53.50 & 35.80 & 44.80 &  \underline{53.00} & \underline{16.00} \\
HRO {~\cite{ref41}} & \underline{31.30} & \underline{34.26} & \underline{37.42} & \textbf{39.89} & \textbf{81.53} & \underline{54.51} &  \underline{37.42} &  \underline{45.16} & 51.78 & \textbf{17.50} \\
MMTF-RU (proposed) & \textbf{38.72} & \textbf{38.86} & \textbf{38.94} & \underline{38.98} & 79.55 & \textbf{55.59} & \textbf{38.94} & \textbf{46.34} & \textbf{55.17} & 15.78 \\ \hline
\end{tabular}
}
\end{table}

\noindent \textbf{Results on EPIC-Kitchens-55.} Table~\ref{tab3} summarizes the performance of various methods on the EPIC-Kitchens-55 action anticipation task. Our MMTF-RU model achieves competitive results, with the highest Top-5 accuracy at most anticipation times, including 38.94\% at $\tau_a = 1s$ and second-best at $\tau_a = 0.5s$ (38.98\%). In Top-5 accuracy at $\tau_a=1s$ for verbs, nouns, MMTF-RU outperforms HRO~\cite{ref41} and Ub-RULSTM~\cite{ref44}. It shows improvements of -1.98\%, +1.08\%, and +1.52\% in Top-5 activity accuracy, and +1.18\%, +2.17\%, and -1.72\% in Mean Top-5 Recall. We also visualize Top-1 action anticipation results at $\tau_a = 0.5s, 1s, 1.5s, 2s$ in Fig.~\ref{fig:sub2}, showing both accurate and incorrect predictions. Model failures often arise from semantic similarities between classes (e.g., \say{pour cereal} vs. \say{take cereal}) or object co-occurrence bias in cluttered scenes (e.g., \say{move bottle} vs. \say{put container}).

\begin{figure}
\centering
\subfloat[]{\includegraphics[width=0.45\textwidth]{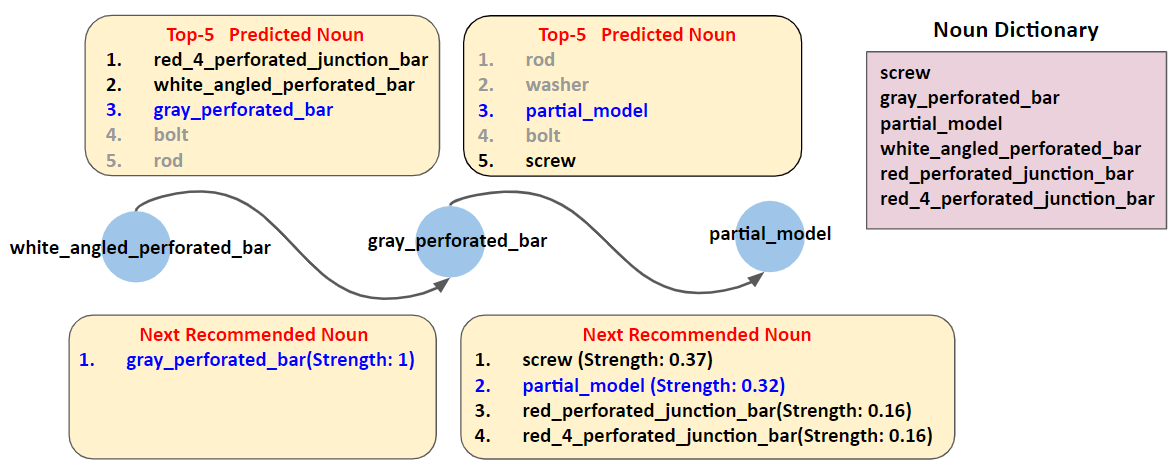}} 
\hfill
\subfloat[]{\includegraphics[width=0.48\textwidth]{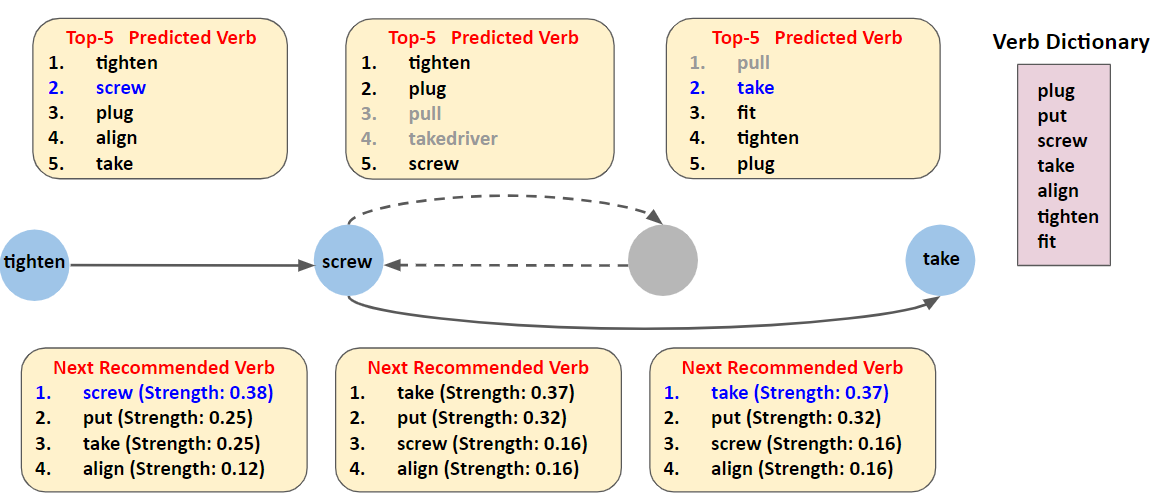}}
\caption{Next (a) noun and (b) verb guidance, based on the alignment between Top-5 MMTF-RU model predictions, the reference graph, and the dictionary set, conducted on the Meccano dataset.}

\label{fig:results_top-5}
\end{figure}

\begin{figure}[!h]
\centering
\begin{tabular}{cc}
    \subfloat[ Meccano - Action Guidance]{\includegraphics[width=0.23\textwidth]{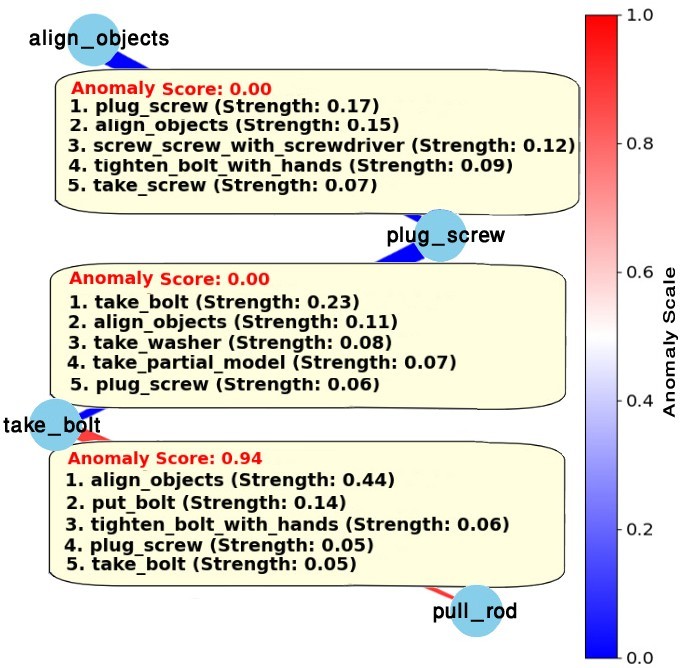}} &
    \subfloat[ Meccano - Noun Guidance]{\includegraphics[width=0.23\textwidth]{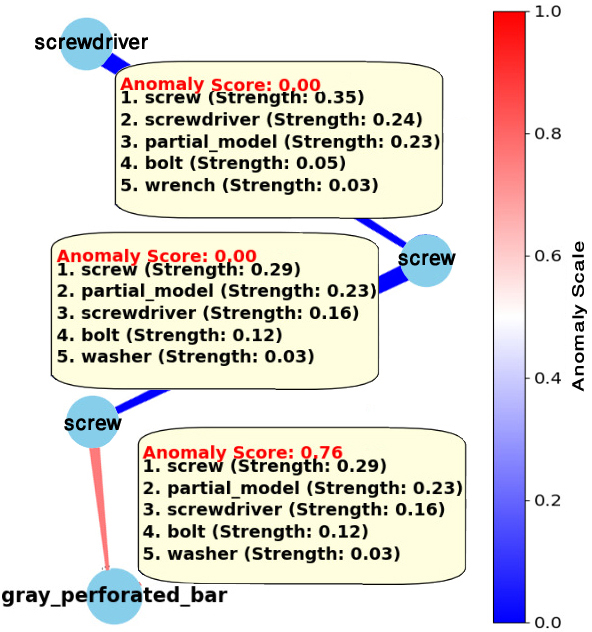}} \\
    \subfloat[ Meccano - Verb Guidance]{\includegraphics[width=0.23\textwidth]{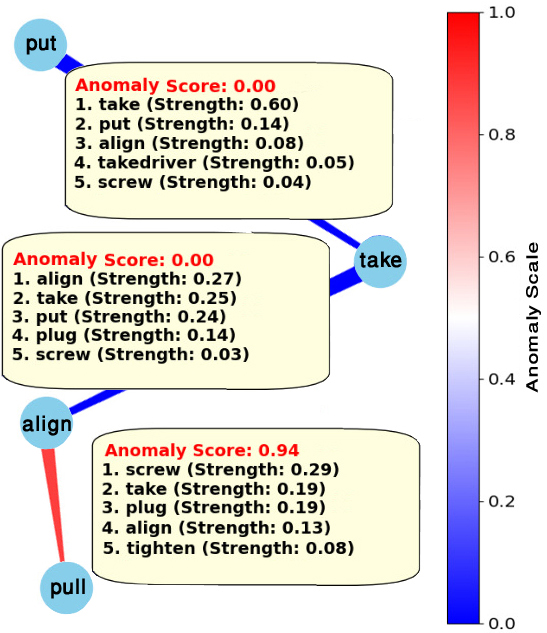}} &
    \subfloat [EK-55 - Action Guidance]{\includegraphics[width=0.23\textwidth]{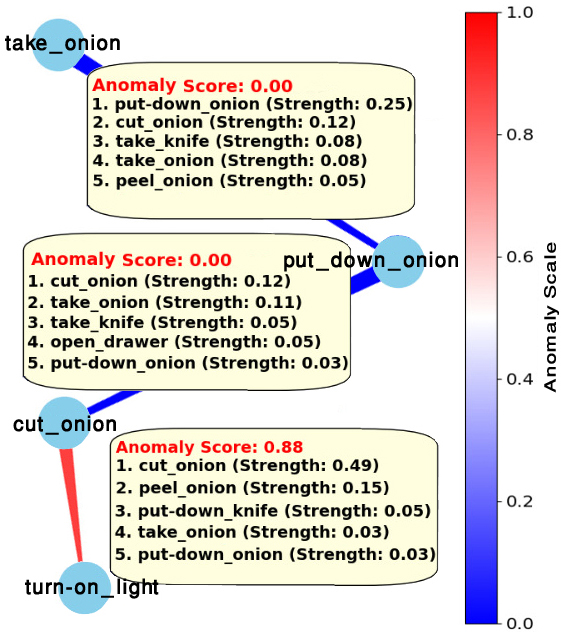}}
\end{tabular}
\caption{Guidance for next actions, nouns, and verbs, along with anomaly scores, based on Top-1 MMTF-RU model predictions and the reference graph.}
\label{fig:anomaly_guidance}
\end{figure}

\subsection{Operator Guidance, Anomaly Prevention, and Task Efficiency Evaluation}

We used the graph-based framework from Algo. \ref{alg:guidance_operator_subgraph}, with results shown in Fig.\ref{fig:results_top-5} (a) and (b). Applied to the Meccano\cite{ref5} dataset, the system guides the operator using MMTF-RU Top-5 predictions at $\tau_{a} = 1s$ and a reference graph for predicting the next verbs and nouns. Blue text indicates agreement between the model and the graph, while grey text highlights discrepancies from the dictionary. Light blue nodes represent normal cases, while grey nodes indicate null cases, prompting action repetition. Starting with the noun \say{white\_angled\_perforated\_bar}, both the graph and MMTF-RU suggest \say{gray\_perforated\_bar} as the next noun, followed by \say{partial\_model}, indicating alignment. For verbs, both suggest \say{screw} as the next verb. In the following step, no verb matches between Top-5 predictions and the graph, prompting the operator to repeat the verb and suggests next verbs. After repetition, the verb \say{take} aligns with both the Top-5 predictions and the graph recommendation. 

\begin{figure}
\centering
\subfloat[]{\includegraphics[width=0.44\textwidth]{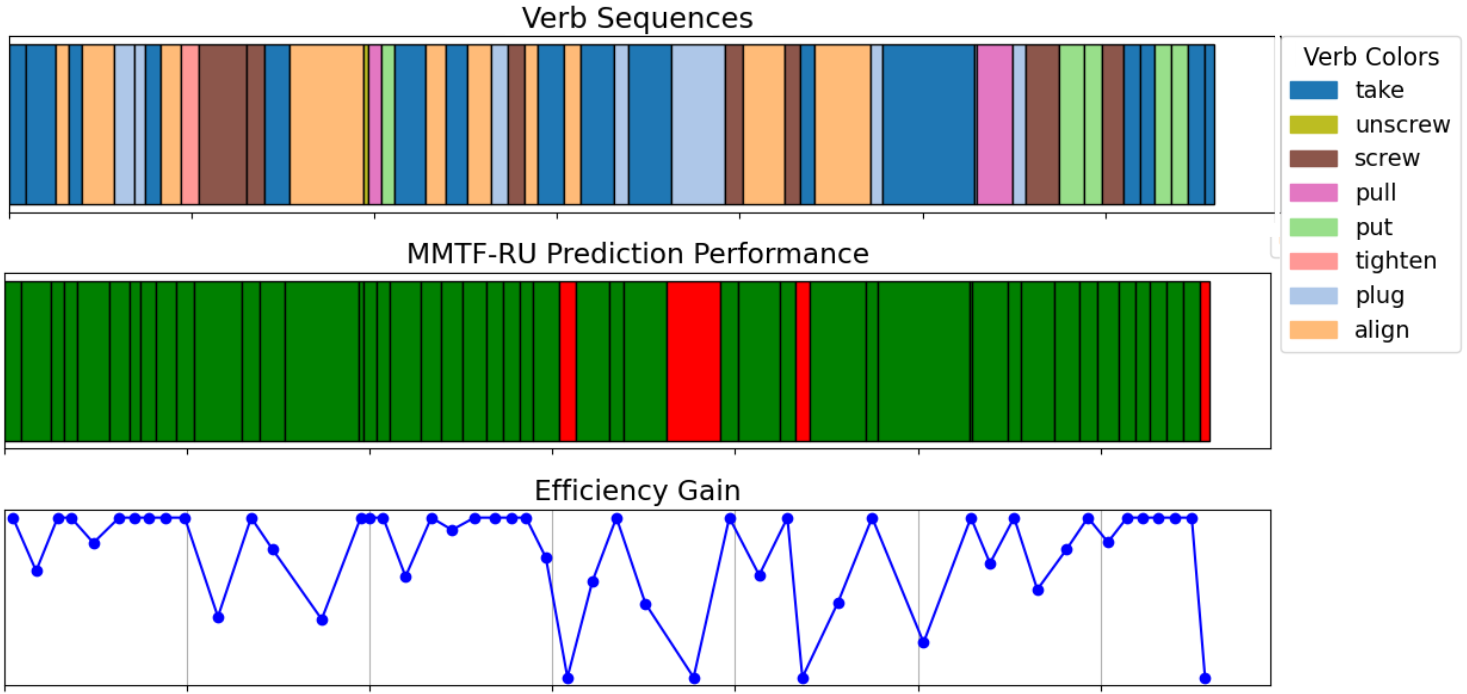}} 
\hfill
\subfloat[]{\includegraphics[width=0.44\textwidth]{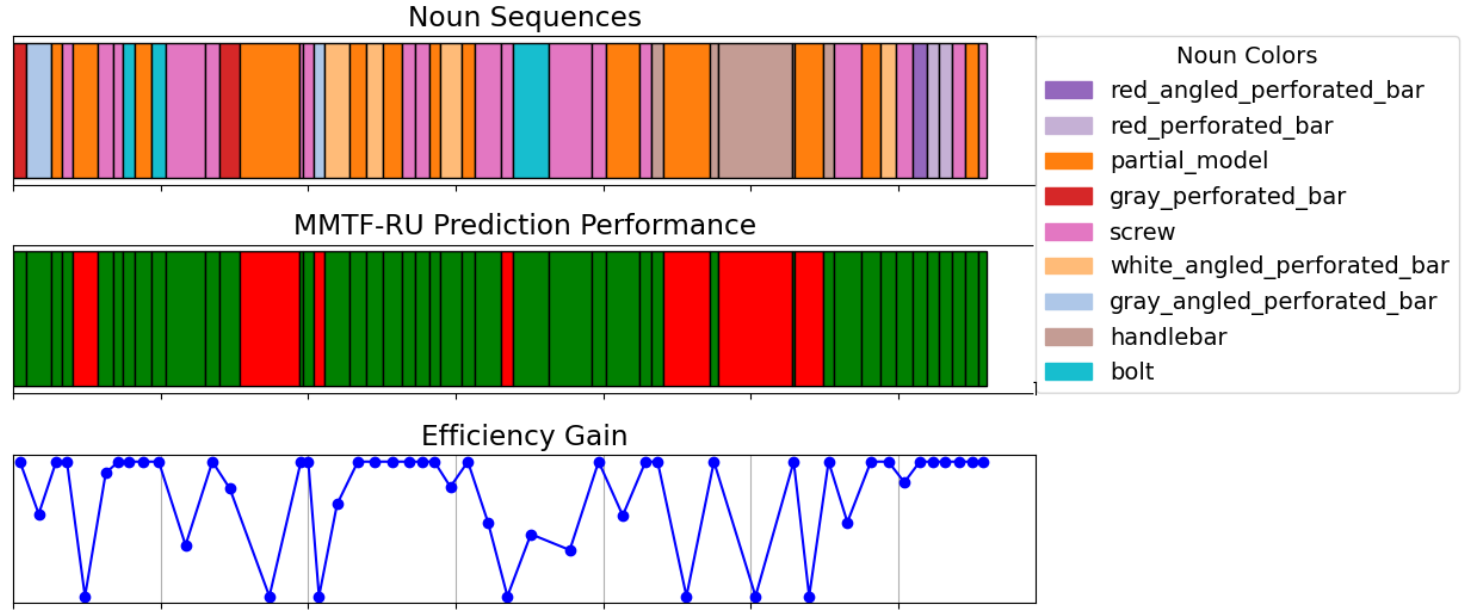}}
\caption{Visualization of operator efficiency for noun and verb sequences from the Meccano dataset.}
\label{seq_compare}
\end{figure}

\begin{figure}
\centering
\subfloat[]{\includegraphics[width=0.23\textwidth]{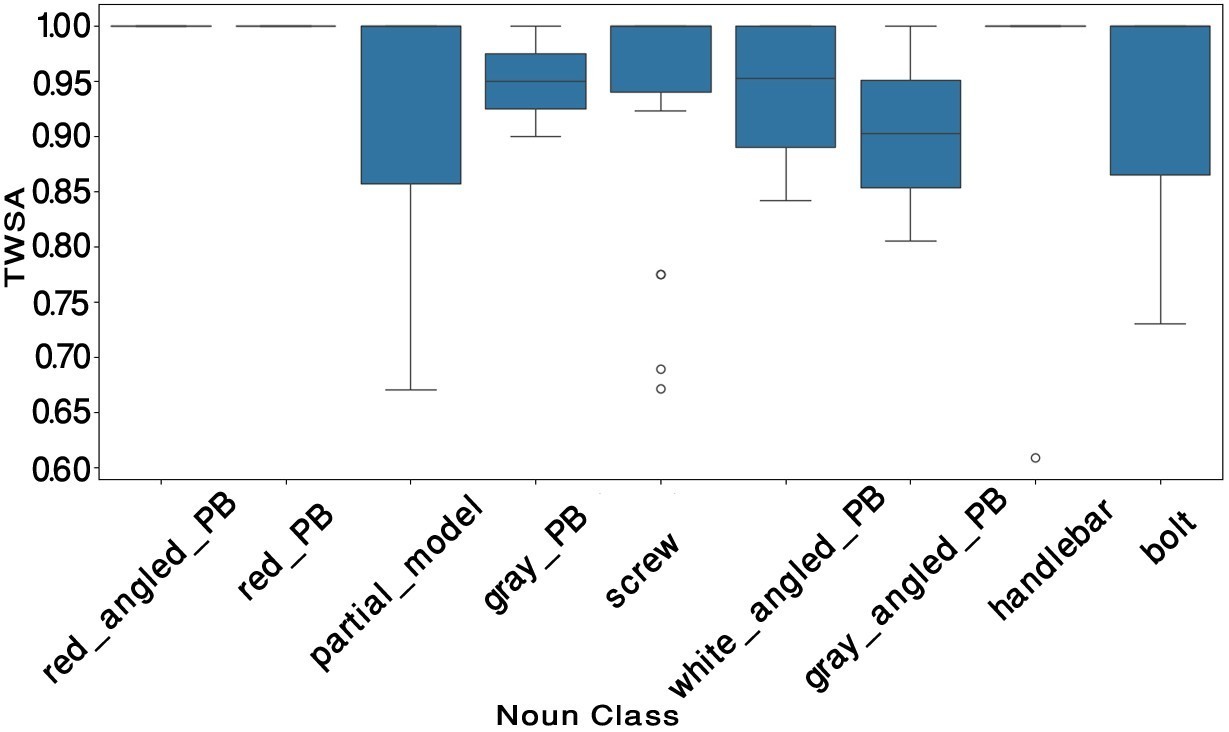}} 
\hfill
\subfloat[]{\includegraphics[width=0.24\textwidth]{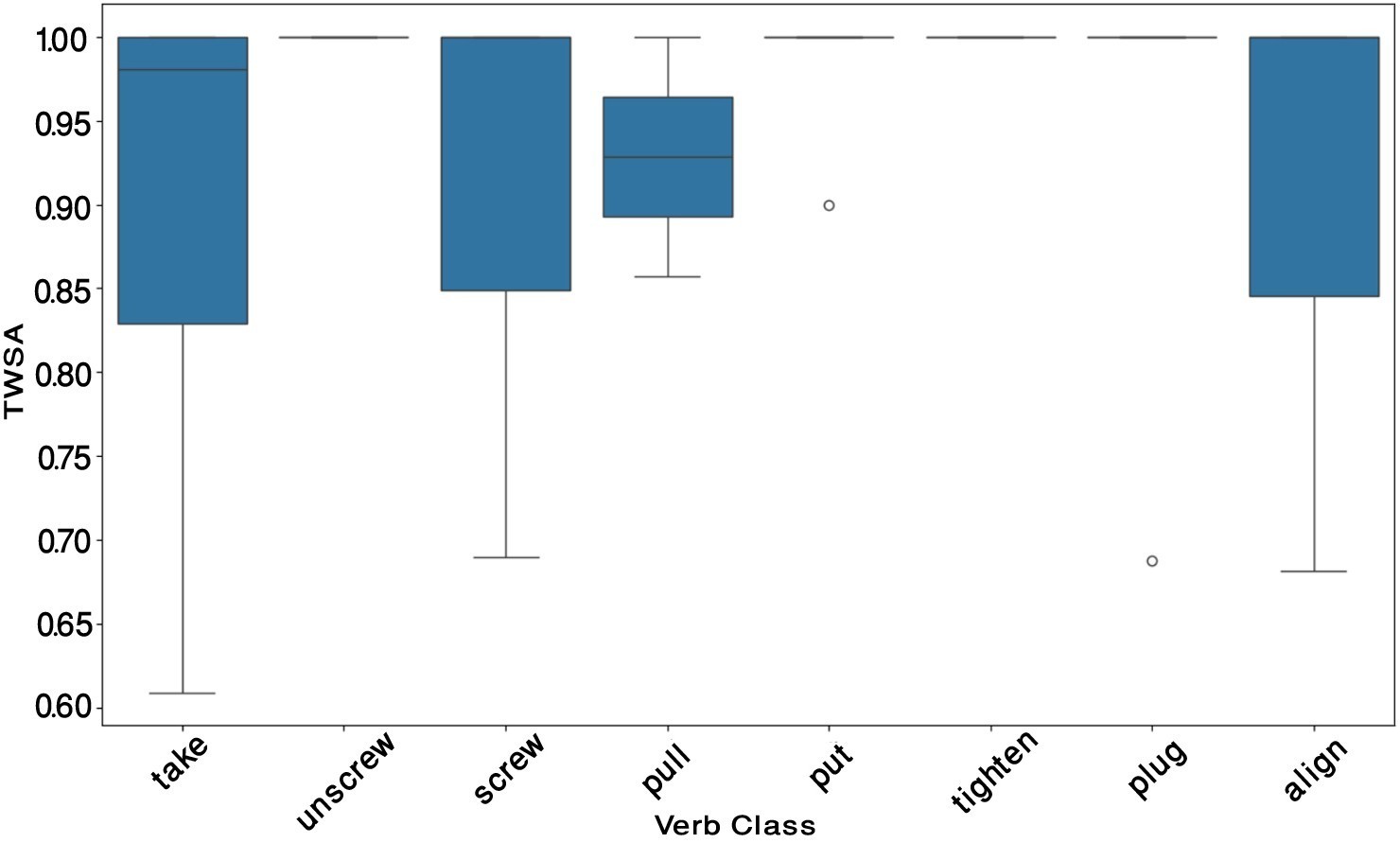}}
\caption{Class-wise TSWA distribution for (a) noun (including PB: perforated\_bar) and (b) verb sequences from the Meccano dataset.}
\label{twsa_combined}
\end{figure}
Next, the MMTF-RU model's predictions at $\tau_{a} =1s$  is integrated with our graph-based anomaly prevention and guidance framework (Algo. \ref{alg:subgraph_and_anomaly}) to analyze task sequences in the Meccano~\cite{ref5} dataset. Fig.~\ref{fig:anomaly_guidance}(a) shows an action sequence where anomaly scores are calculated for task transitions. Starting with \say{align\_objects} and \say{plug\_screw}, the sequence progresses as expected with zero anomaly scores. The transition from \say{plug\_screw} to \say{take\_bolt} is also recommended. However, transitioning from \say{take\_bolt} to \say{pull\_rod} triggers a high anomaly score of 0.94, indicating a deviation from the expected flow. Alternatives such as \say{align\_objects} (strength: 0.44) and \say{put\_bolt} (strength: 0.14) align more closely with the expected sequence, identifying \say{pull\_rod} as a source of inefficiency. The color gradient, from blue to red, represents the severity of anomalies. Fig.~\ref{fig:anomaly_guidance}(b) highlights noun sequence analysis, where the transition  \say{screw} to \say{gray\_perforated\_bar} results in an anomaly score of 0.76. Fig.~\ref{fig:anomaly_guidance}(c) shows verb sequence analysis, with a significant anomaly score of 0.94 during the transition from \say{align} to \say{pull}. Finally, Fig.~\ref{fig:anomaly_guidance}(d) demonstrates the method's validation on the EPIC-Kitchens-55~\cite{ref6} dataset, where a high anomaly score of 0.88 occurs during the transition from \say{cut\_onion} to \say{turn-on\_light}. The guidance model effectively identifies potential issues, providing suggested corrections based on transition strengths.

Fig.~\ref{seq_compare} illustrates the sequential assembly tasks for noun and verb classes using 50 test samples from the Meccano dataset. The width of the color-coded sequences represents the time taken by the operator for each task. Below, the Top-5 MMTF-RU predictions are displayed (Red: incorrect, Green: correct). Efficiency gain is visualized through the relative scores calculated for each step using Eq.~\ref{eq10}. Fig.~\ref{twsa_combined}(a) shows the class-wise distribution of TWSA for noun classes. \say{red\_perforated\_bar}, \say{red\_angled\_perforated\_bar}, and \say{handlebar} exhibit high and consistent TWSA values, indicating efficient performance with minimal deviation from the optimal sequence. Conversely, classes like \say{partial\_model}, \say{gray\_perforated\_bar}, and \say{bolt} show greater variability and lower TWSA, suggesting inefficiencies. Outliers in the \say{screw} class highlight areas for improvement. For verb classes (Fig.~\ref{twsa_combined}(b)), actions like \say{unscrew}, \say{put}, and \say{tighten} maintain high TWSA, while \say{take}, \say{screw}, and \say{align} display broader TWSA ranges, indicating inconsistencies. Outliers in \say{pull} and \say{plug} suggest further optimization potential. The overall TWSA for verb and noun classes are 0.86 and 0.84, respectively, showing a high level of sequence accuracy with room for improvement. This analysis identifies specific classes for performance enhancement, aiding in optimizing assembly tasks and efficiency.

\subsection{Ablation Study}

\begin{table}[b]
\centering
\caption{Comparison of various modality inputs to the CMFB and analysis of the GRU's hidden layer input $(h_0)$ on the Meccano~\cite{ref5} and EPIC-Kitchens-55~\cite{ref6} datasets. The symbol $\bigoplus$ denotes the concatenation of available modalities from each dataset.}

\label{tab4} 
\resizebox{0.46\textwidth}{!}{%
\begin{tabular}{clllll}
\hline
\multicolumn{6}{c}{\textbf{Top-1/Top-5 @1s}}                                                                                                           \\ \hline
\multicolumn{1}{c|}{\textbf{Dataset}}                  & \multicolumn{1}{c|}{$\bm{f}_{1}^{0}$} & \multicolumn{1}{c|}{$\bm{f}_{2}^{0}$} & \multicolumn{1}{c|}{$\bm{f}_{3}^{0}$} & \multicolumn{1}{c|}{$\textbf{h}_0 \gets 0$} & $\text{L}_{\text{hid}}\left(\bigoplus_{m \in \{1, 2, 3\}} \bm{f}_{m}^{0}\right)$  \\ \hline

\multicolumn{1}{c|}{\multirow{3}{*}{Meccano~\cite{ref5}}} & \multicolumn{1}{c|}{\checkmark} & \multicolumn{1}{c|}{} & \multicolumn{1}{c|}{} & \multicolumn{1}{c|}{28.10/61.68} & \multicolumn{1}{c}{28.19/61.81} \\
\multicolumn{1}{c|}{}  & \multicolumn{1}{c|}{\checkmark} & \multicolumn{1}{c|}{\checkmark} & \multicolumn{1}{c|}{} & \multicolumn{1}{c|}{27.92/61.19} & \multicolumn{1}{c}{28.12/62.94} \\
\multicolumn{1}{c|}{}                         & \multicolumn{1}{c|}{}                         & \multicolumn{1}{c|}{\checkmark} & \multicolumn{1}{c|}{\checkmark} & \multicolumn{1}{c|}{27.88/61.01} & \multicolumn{1}{c}{28.06/62.59}    \\

\multicolumn{1}{c|}{}                        & \multicolumn{1}{c|}{\checkmark}                         & \multicolumn{1}{c|}{} & \multicolumn{1}{c|}{\checkmark} & \multicolumn{1}{c|}{27.53/61.13} & \multicolumn{1}{c}{28.28/63.23}  \\

\multicolumn{1}{c|}{}                        & \multicolumn{1}{c|}{\checkmark}                         & \multicolumn{1}{c|}{\checkmark} & \multicolumn{1}{c|}{\checkmark} & \multicolumn{1}{c|}{28.20/63.40} & \multicolumn{1}{c}{\textbf{29.75/64.46}}    \\ \hline

\multicolumn{1}{c|}{\multirow{3}{*}{EPIC-Kitchens-55~\cite{ref6}}}  & \multicolumn{1}{c|}{\checkmark} & \multicolumn{1}{c|}{ } & \multicolumn{1}{c|}{} & \multicolumn{1}{c|}{15.55/33.30 } & \multicolumn{1}{c}{15.65/33.36} \\
\multicolumn{1}{c|}{}  & \multicolumn{1}{c|}{\checkmark} & \multicolumn{1}{c|}{\checkmark} & \multicolumn{1}{c|}{} & \multicolumn{1}{c|}{15.60/33.18} & \multicolumn{1}{c}{15.89/33.39} \\
\multicolumn{1}{c|}{}                         & \multicolumn{1}{c|}{}                         & \multicolumn{1}{c|}{\checkmark} & \multicolumn{1}{c|}{\checkmark} & \multicolumn{1}{c|}{16.66/36.46} & \multicolumn{1}{c}{17.01/37.09}    \\

\multicolumn{1}{c|}{}                        & \multicolumn{1}{c|}{\checkmark}                         & \multicolumn{1}{c|}{} & \multicolumn{1}{c|}{\checkmark} & \multicolumn{1}{c|}{17.26/37.49} & \multicolumn{1}{c}{17.80/38.33}  \\

\multicolumn{1}{c|}{}                        & \multicolumn{1}{c|}{\checkmark}                         & \multicolumn{1}{c|}{\checkmark} & \multicolumn{1}{c|}{\checkmark} & \multicolumn{1}{c|}{18.20/38.09} & \multicolumn{1}{c}{ \textbf{18.44/38.94}}    \\ \hline

\end{tabular} %
}
\end{table} 

\begin{figure}
\centering
\subfloat[]{\includegraphics[width=0.44\textwidth]{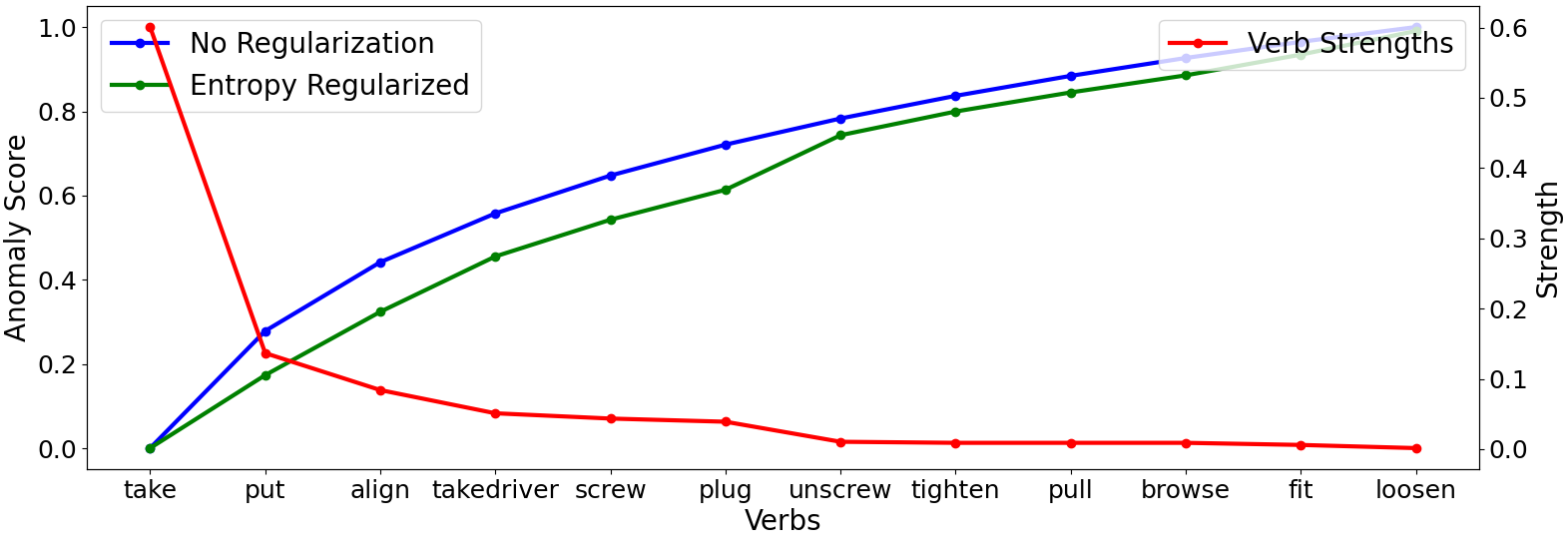}}  \\
\subfloat[]{\includegraphics[width=0.44\textwidth]{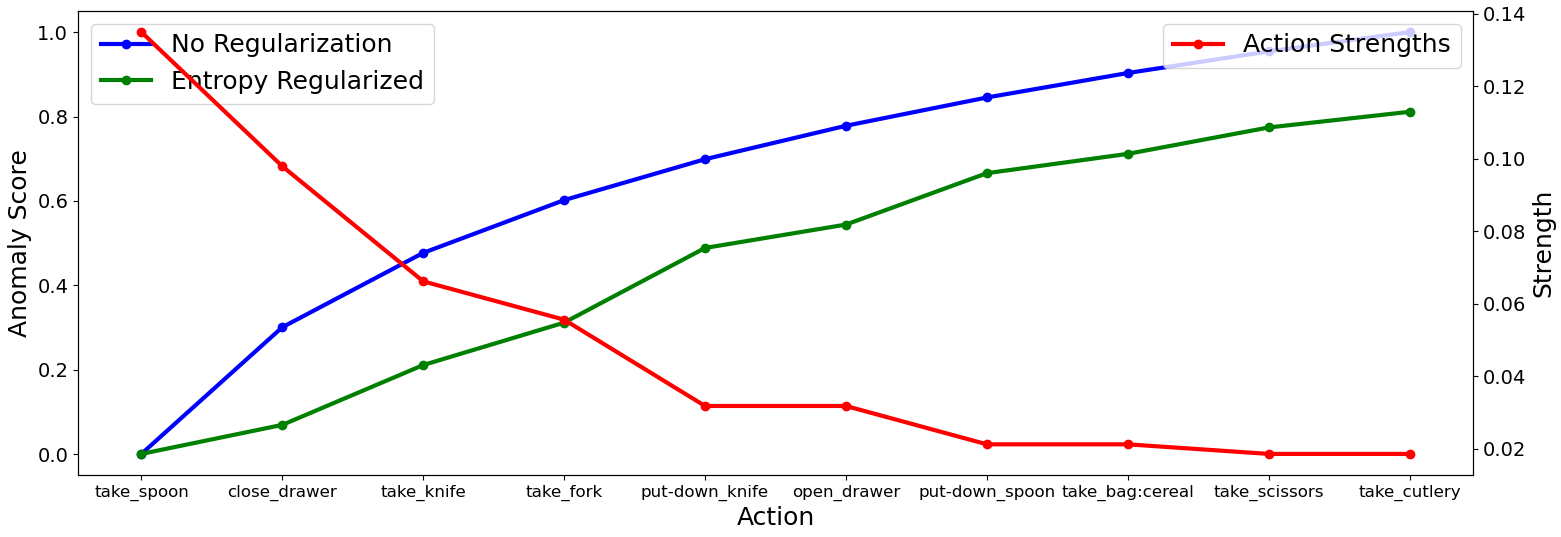}}
\caption{Anomaly score comparison with and without action certainty on (a) Meccano dataset and (b) EK-55 dataset.}
\label{abln_combined}
\end{figure}
Table~\ref{tab4} presents the impact of input modality pairs in the CMFB and the influence of initializing the decoder's GRU block hidden layer $h_0$. Focusing on the Meccano~\cite{ref5} and EPIC-Kitchens-55~\cite{ref6} datasets, we assess the performance of combining input modality features $\bm{f}_{1}^{0}$, $\bm{f}_{2}^{0}$, and$\bm{f}_{3}^{0}$, representing object/RGB, hand/flow, and gaze/object features for Meccano/EPIC-Kitchens-55 in the CMFB through ablative experiments. Results demonstrate that incorporating all modalities in the CMFB enhances model performance in both cases of GRU's hidden layer initialization, enhancing the model's discriminative ability through pair-wise fusion of complementary features. We also conducted experiments involving guided $\textbf{h}_0 $ initialization, as described in Equation \ref{eq2}. The method entails integrating object/RGB, hand/flow, and gaze/object features through concatenation, followed by a linear transformation before feeding them as $\textbf{h}_0 $. The results demonstrate a 1.55\% and 0.24\% improvement in Top-1, and a 1.06\% and 0.85\% improvement in Top-5 action anticipation accuracy compared to using $\textbf{h}_0  = 0$ initialization for the Meccano~\cite{ref5} and EPIC-Kitchens-55~\cite{ref6} datasets, respectively.  In another ablation study, we assess the impact of incorporating  action certainty into the anomaly score calculation in Algo. \ref{alg:subgraph_and_anomaly}. This integration produces more refined anomaly scores, reducing penalties for expected actions while sharply penalizing significant deviations. Fig.~\ref{abln_combined} shows that certainty-based scaling, driven by entropy, improves anomaly detection by adjusting the anomaly score. This method avoids over-penalizing minor deviations while emphasizing significant ones, using prediction uncertainty to apply appropriate penalties and better detect true anomalies.

\section{Conclusion}
\label{CN}

Real-time assembly monitoring is crucial for preventing errors and maintaining product quality. This work introduces the MMTF-RU model for egocentric activity anticipation, paired with the OAMU framework to predict operator actions and address deviations using Top-1/Top-5 predictions and a reference graph. Our method achieves state-of-the-art results on the Meccano industrial dataset and competitive performance on EPIC-Kitchens-55, highlighting its robustness. Despite variability in assembly tasks, the integrated framework provides accurate next-step guidance and anomaly prevention, enhancing decision-making and optimizing task flow. To further enhance efficiency, we propose the TWSA metric, which identifies bottlenecks and ensures smooth task execution, leading to streamlined and error-free processes. Future work will incorporate operator feedback to enhance adaptability and validate the framework in complex industrial settings, addressing task allocation and scheduling challenges.

\section*{Acknowledgment}
The authors express their gratitude to the Director of CSIR-CEERI for encouraging AI-related research activities. This study was conducted under the \say{Resource Constrained AI} project, funded by the Ministry of Electronics and Information Technology (MeitY), India. Naval Kishore Mehta thanks CSIR-HRDG for the CSIR-SRF-Direct fellowship support.

\bibliographystyle{IEEEtran}  
\bibliography{bib}

\begin{IEEEbiography}[{\includegraphics[width=1in,height=1.25in,clip,keepaspectratio]{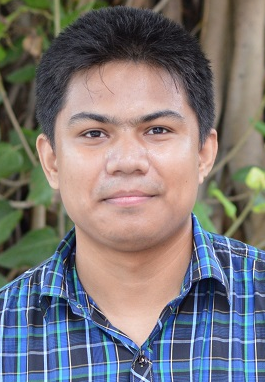}}] {Naval Kishore Mehta} is currently an Integrated Dual Degree Ph.D. (IDDP) Program student in the Academy of Scientific and Innovative Research (AcSIR), at Advanced Information Technologies Group, CSIR-Central Electronics Engineering Research Institute (CSIR-CEERI) Campus, Pilani, India. His research focuses on human action recognition and anticipation, as well as exploring deep learning applications for industrial use cases.
\end{IEEEbiography}

\begin{IEEEbiography}[{\includegraphics[width=1in,height=1.25in,clip,keepaspectratio]{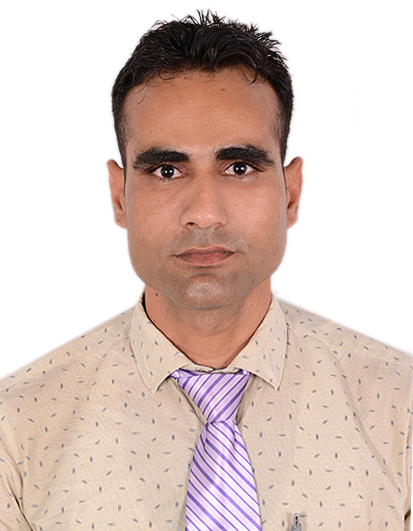}}] {Arvind} received the B.Tech degree in Information Technology from Sobhasaria Engineering College, Sikar in 2013, and an M.Tech in Data Science and Engineering from BITS Pilani. With over a decade of industrial experience, he has worked in various sectors across India, Saudi Arabia, and UAE. His recent roles include positions at the University of Najran in Saudi Arabia, Ministry of Education-UAE and Expo 2020 Dubai, as well as CSIR-CEERI Pilani. Currently pursuing a Ph.D. at AcSIR, CSIR-CEERI Pilani, his research interests focus on developing AI-enabled computer vision techniques for industrial environments.
\end{IEEEbiography}

\begin{IEEEbiography}[{\includegraphics[width=1in,height=1.25in,clip,keepaspectratio]{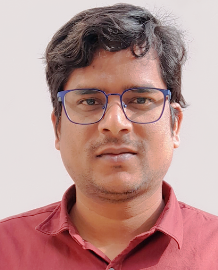}}] {Shyam Sunder Prasad} received the B.Tech degree in electronics and telecommunication from Biju Patnaik University of Technology, Rourkela, Odisha, India, in 2012, and the M.Tech in advanced electronic systems from the Academy of Scientific and Innovative Research (AcSIR), Ghaziabad, India, in 2017. He is currently pursuing the Ph.D. degree from Academy of Scientific and Innovative Research (AcSIR), at Advanced Information Technologies Group, CSIR-Central Electronics Engineering Research Institute (CSIR-CEERI) Campus, Pilani, India.
His research interests include face anti-spoofing solutions using deep learning techniques, design of face biometric systems, and computer vision application based on spiking neural networks.
\end{IEEEbiography}

\begin{IEEEbiography}[{\includegraphics[width=1in,height=1.25in,clip,keepaspectratio]{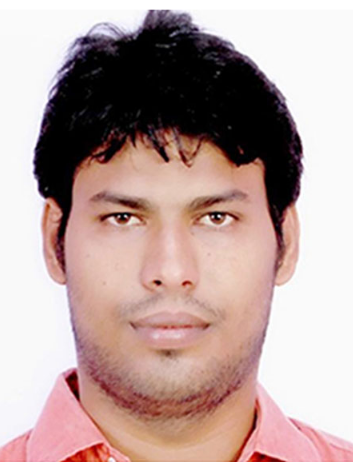}}] {Sumeet Saurav} received the M.Tech degree from the Advanced Semiconductor Electronics and Ph.D. degree from the Academy of Scientific and Innovative Research (AcSIR), Ghaziabad, India, in 2014 and 2022, respectively.
He is working as a Senior Scientist with the Advanced Information Technologies Group at CSIR-Central Electronics Engineering Research Institute (CSIR-CEERI), Pilani, India. He joined CSIR-CEERI as Quick Hire Fellow (QHF), in 2012. His research interests include Computer Vision, Machine Learning, Deep Learning Architecture design for Computer Vision Applications, and Embedded Real-Time Implementation of Computer Vision Algorithms.
\end{IEEEbiography}

\begin{IEEEbiography}[{\includegraphics[width=1in,height=1.25in,clip,keepaspectratio]{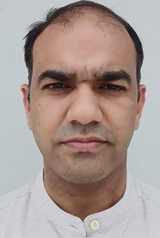}}]{Sanjay Singh} received his B.Sc. in Electronics and Computer Science in 2003, M.Sc. in Electronics in 2005, M.Tech. in Microelectronics and VLSI Design in 2007, and Ph.D. in VLSI Design for Computer Vision Applications in 2015. 

He joined CSIR-Central Electronics Engineering Research Institute (CSIR-CEERI), Pilani, as a Scientist Fellow in 2009. He currently serves as the Principal Scientist and Head of the Advanced Information Technologies Group at CSIR-CEERI, Pilani, India. Additionally, he holds the position of Associate Professor (Engineering Sciences) at the Academy of Scientific and Innovative Research (AcSIR), Ghaziabad, India. His research interests include Computer Vision, Machine Learning, Artificial Intelligence, VLSI Architectures, and FPGA Prototyping.

\end{IEEEbiography} 

\end{document}